\documentclass[11pt]{article}

\usepackage{microtype}
\usepackage{graphicx}
\usepackage{subcaption}
\usepackage{booktabs}
\usepackage{longtable}
\usepackage{textcomp}
\usepackage{microsoft-tech-report}
\usepackage{hyperref}
\usepackage{amsmath}
\usepackage{amssymb}
\usepackage{mathtools}
\usepackage{amsthm}
\usepackage{bm}
\usepackage{enumitem}
\usepackage[capitalize,noabbrev]{cleveref}
\usepackage{xspace}
\usepackage{titletoc}
\usepackage{placeins}  
\usepackage{afterpage} 
\usepackage{wrapfig}   

\renewcommand{\topfraction}{0.9}      
\renewcommand{\bottomfraction}{0.7}
\renewcommand{\textfraction}{0.07}    
\renewcommand{\floatpagefraction}{0.7}
\usepackage{tcolorbox} 
\usepackage{multirow}
\usepackage{listings}  
\tcbuselibrary{theorems,skins,breakable,listings}

\definecolor{promptbg}{HTML}{F7F7F8}
\definecolor{promptborder}{HTML}{5B5B5B}
\newtcolorbox{promptbox}[1][]{
  colback=promptbg, colframe=promptborder, fonttitle=\bfseries\sffamily,
  breakable, enhanced, arc=3pt, boxrule=0.8pt,
  left=6pt, right=6pt, top=5pt, bottom=5pt,
  before skip=6pt, after skip=6pt,
  title={#1}
}

\lstdefinestyle{json}{
  basicstyle=\fontsize{8}{9.6}\selectfont\ttfamily,
  breaklines=true,
  breakatwhitespace=true,
  showstringspaces=false,
  columns=fullflexible,
  keepspaces=true,
  upquote=true,
  literate={"}{\textquotedbl}1,
}

\newif\ifappendixtoc
\appendixtoctrue   

\definecolor{color_blue}{HTML}{E7EFFA}
\definecolor{color_green}{HTML}{E6F8E0}
\definecolor{color_gray}{HTML}{ECECEC}
\definecolor{pearDark}{HTML}{2980B9}
\definecolor{theoremblue}{HTML}{EBF5FB}
\definecolor{theoremborder}{HTML}{2980B9}
\definecolor{propgreen}{HTML}{EAFAF1}
\definecolor{propborder}{HTML}{27AE60}
\definecolor{defyellow}{HTML}{FEF9E7}
\definecolor{defborder}{HTML}{F39C12}
\definecolor{remarkgray}{HTML}{F2F3F4}
\definecolor{remarkborder}{HTML}{7F8C8D}

\hypersetup{
  colorlinks=true,
  linkcolor=msftblue,
  citecolor=msftblue,
  urlcolor=msftblue,
  pdftitle={AesCode: Aesthetic Code Generation with Decoupled Cross-Modal Rewards}
}

\newtcbtheorem[auto counter]{theorem}{Theorem}{
  colback=theoremblue, colframe=theoremborder, fonttitle=\bfseries,
  breakable, enhanced, sharp corners, boxrule=0.6pt, left=6pt, right=6pt, top=6pt, bottom=6pt
}{thm}

\newtcbtheorem[use counter from=theorem]{proposition}{Proposition}{
  colback=propgreen, colframe=propborder, fonttitle=\bfseries,
  breakable, enhanced, sharp corners, boxrule=0.6pt, left=6pt, right=6pt, top=6pt, bottom=6pt
}{prop}

\newtcbtheorem[use counter from=theorem]{lemma}{Lemma}{
  colback=theoremblue, colframe=theoremborder, fonttitle=\bfseries,
  breakable, enhanced, sharp corners, boxrule=0.6pt, left=6pt, right=6pt, top=6pt, bottom=6pt
}{lem}

\newtcbtheorem[use counter from=theorem]{corollary}{Corollary}{
  colback=propgreen, colframe=propborder, fonttitle=\bfseries,
  breakable, enhanced, sharp corners, boxrule=0.6pt, left=6pt, right=6pt, top=6pt, bottom=6pt
}{cor}

\newtcbtheorem[use counter from=theorem]{definition}{Definition}{
  colback=defyellow, colframe=defborder, fonttitle=\bfseries,
  breakable, enhanced, sharp corners, boxrule=0.6pt, left=6pt, right=6pt, top=6pt, bottom=6pt
}{def}

\newtcbtheorem[use counter from=theorem]{assumption}{Assumption}{
  colback=defyellow, colframe=defborder, fonttitle=\bfseries,
  breakable, enhanced, sharp corners, boxrule=0.6pt, left=6pt, right=6pt, top=6pt, bottom=6pt
}{asm}

\newtcbtheorem[use counter from=theorem]{remark}{Remark}{
  colback=remarkgray, colframe=remarkborder, fonttitle=\bfseries,
  breakable, enhanced, sharp corners, boxrule=0.6pt, left=6pt, right=6pt, top=6pt, bottom=6pt
}{rem}

\crefname{tcb@cnt@theorem}{Theorem}{Theorems}
\crefname{tcb@cnt@proposition}{Proposition}{Propositions}
\crefname{tcb@cnt@lemma}{Lemma}{Lemmas}
\crefname{tcb@cnt@corollary}{Corollary}{Corollaries}
\crefname{tcb@cnt@definition}{Definition}{Definitions}
\crefname{tcb@cnt@assumption}{Assumption}{Assumptions}
\crefname{tcb@cnt@remark}{Remark}{Remarks}

\newcommand{\method}{AesCode\xspace}   
\techreportshorttitle{AesCode}

\newcommand{\corrmark}{\dagger}
\newcommand{\corremail}[1]{\href{mailto:#1}{#1}}
\newcommand{\corremailpair}[3]{%
  \{\href{mailto:#1@#3}{#1}, \href{mailto:#2@#3}{#2}\}\href{mailto:#1@#3}{@#3}}
\newcommand{\corremails}{\corremail{yifanyang@microsoft.com},
  \corremailpair{leandros.tassiulas}{rex.ying}{yale.edu}}
\newcommand{\mainresultstablefont}{\fontsize{8.8}{10.4}\selectfont}
\newcommand{\mainresultstabcolsep}{6pt}
\newcommand{\mainresultscontentheader}{\fontsize{8.2}{9.8}\selectfont\textbf{Content}}
\newcommand{\microsoftanalysisfloats}{}

\begin{document}
\thispagestyle{empty}

\noindent
\begin{minipage}[c]{0.5\linewidth}
\raggedright
\raisebox{-0.5\height}{\msftbrandmark}
\end{minipage}
\begin{minipage}[c]{0.49\linewidth}
\raggedleft
{\msftdatefont\small\color{msftgray}Oct 2026}    
\end{minipage}\par
\vspace{0.35em}
\noindent{\color{msftline}\rule{\linewidth}{0.8pt}\par}

\vspace{0.85em}
\begin{center}
\setlength{\parskip}{0pt}
\linespread{1}%
{{\msfttitlefont\fontsize{19}{21}\selectfont\color{msftdark}
AesCode: Aesthetic Code Generation with Decoupled Cross-Modal Rewards\par}}
\vspace{1.2em}

{\fontsize{10.4pt}{14.2pt}\selectfont\rmfamily\color{msftdark}
Xin Gao$^{2,*}$ \quad Yifan Yang$^{1,\corrmark}$ \quad Xin Li$^{3}$ \quad
Pan Wang$^{4}$ \quad Yan Li$^{5}$ \quad Muzhao Tian$^{6}$\\
Mingxi Cheng$^{1}$ \quad Bei Liu$^{1}$ \quad Kai Qiu$^{1}$ \quad
Qi Dai$^{1}$ \quad Dong Chen$^{1}$ \quad Yue Dong$^{1}$\\
Leandros Tassiulas$^{2,\corrmark}$ \quad Rex Ying$^{2,\corrmark}$ \quad
Ji Li$^{1}$ \quad Chong Luo$^{1}$\par
}
\vspace{0.25cm}

{\fontsize{8.8pt}{11.8pt}\selectfont\rmfamily\color{msftgray}
\textcolor{msftblue}{$^{1}$}\,Microsoft \hspace{1.35em}
\textcolor{msftblue}{$^{2}$}\,Yale University \hspace{1.35em}
\textcolor{msftblue}{$^{3}$}\,Nanyang Technological University\\
\textcolor{msftblue}{$^{4}$}\,University of Pittsburgh \hspace{1.35em}
\textcolor{msftblue}{$^{5}$}\,Shanghai Jiao Tong University \hspace{1.35em}
\textcolor{msftblue}{$^{6}$}\,Fudan University\par
}
\end{center}

\vspace{0.45em}
\begin{msfttitlebox}
\setlength{\parindent}{0cm}
\setlength{\parskip}{0.14cm}

{\fontsize{10.2pt}{13.2pt}\selectfont
\noindent\makebox[\linewidth][c]{\begin{minipage}{\dimexpr\linewidth-16pt\relax}
\setlength{\parfillskip}{0.16\linewidth plus 1fil}
Generating information-rich visual artifacts such as slides, posters, and dashboards as code is appealing because the outputs are structured, editable, and verifiable. Yet code models cannot see how their choices of layout and color come together on the canvas, often limiting spatial planning and aesthetic quality. 
Image generators excel at visual design but can misrender requested content, making their outputs unreliable targets for direct reproduction. We therefore introduce \method, which generates code grounded in the content prompt while using an image generated from the same prompt as an aesthetic guidance.
To do so, AesCode represents the target as a design graph that describes the whole canvas and makes each property attributable. From this graph, we derive complementary cross-modal rewards: deterministic verifiers score properties parsable from code and rendering, while a VLM judge assesses the non-parsable ones with a sample-specific Visual Graph Rubric (VGR) tied to graph elements and relations. 
We optimize these decoupled rewards through reinforcement learning with per-channel normalization to jointly improve spatial planning, aesthetic quality, and verifiable correctness.
To evaluate this setting, we build a benchmark that scores verifiable correctness and visual quality separately, with visual scores highly consistent with human judgments. We find that reference images substantially improve the visual quality of code generated by existing models, and ablations show that \method benefits from references both as model input and as a source of visual rewards. Relative to same-scale \mbox{Qwen3-VL} baselines given the same references, \mbox{\method-8B} and \mbox{\method-32B} improve visual quality by 31.1 and 22.4 percentage points, reaching scores comparable to those of closed-source frontier models such as \mbox{GPT-5.5} and Claude Opus~4.8.

\end{minipage}}\par}

\vspace{0.22cm}
{\leftskip=8pt\rightskip=8pt
\noindent{\color{msftline}\rule{\dimexpr\linewidth-16pt\relax}{0.4pt}}\par
\vspace{0.16cm}
{\setlength{\parskip}{0.06cm}\small
{\msftmetalabel{Project Page}\href{https://microsoft.github.io/AesCode/}{https://microsoft.github.io/AesCode/}\par}
{\msftmetalabel{Correspondence}\corremails\par}
}
\vspace{0.08cm}
{\footnotesize\rmfamily\itshape\color{msftgray}
\ifx\corremails\empty\else$^\dagger$ Corresponding authors.\hspace{1.1em}\fi
$^*$ Work was done during an internship at Microsoft Research Asia.\par
}}
\end{msfttitlebox}
\suppressfloats[t]  

\begin{figure}[t]
  \centering
  \includegraphics[width=0.98\textwidth]{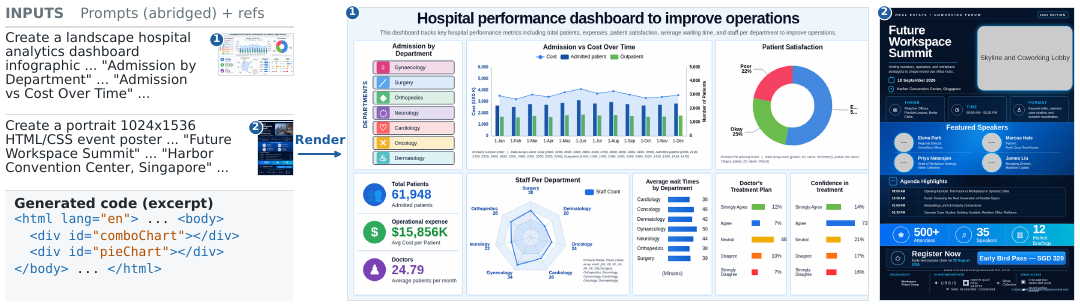}
  \caption{\textbf{Information-rich infographics generated as code.} Prompts specify content, while image-generator references offer visual design guidance. The examples show a landscape dashboard and portrait poster rendered from generated HTML/CSS.}
  \label{fig:teaser}
\end{figure}

\fontsize{10.5pt}{13.4pt}\selectfont
\setlength{\parskip}{6pt}
\setlength{\textfloatsep}{12pt plus 2pt minus 4pt}
\setlength{\floatsep}{10pt plus 2pt minus 2pt}
\setlength{\intextsep}{12pt plus 2pt minus 2pt}

\ifappendixtoc
  \addtocontents{toc}{\protect\setcounter{tocdepth}{-1}}
\fi

\section{Introduction}
\label{sec:introduction}

Information-rich visual artifacts such as slides, posters, and dashboards pack dense content into a single canvas~\citep{pang2025paper2poster, zheng2025pptagent}. Generating them as code is highly desirable because the output remains structured, editable, and verifiable. However, code models suffer from a fundamental limitation: they cannot ``see'' the canvas. Spatial layout, hierarchy, and color are declared line by line, resulting in outputs that satisfy content constraints but often appear visually dated. In contrast, image generators excel at composing visually striking pages~\citep{esser2024scaling, gao2025postermaker}, yet frequently misrender text, numbers, and logical relations, producing raster outputs that cannot be edited or verified~\citep{liu2023character, tang2026igenbench}.

These complementary strengths motivate a novel setting where an image generator provides an \emph{aesthetic guidance} for the code model. Specifically, a reference image generated from the same prompt serves as a conditional input to supply layout and style inspiration. While existing image-to-code methods~\citep{beltramelli2018pix2code, si2025design2code, han2023chartllama} also use images as input, they treat them as absolute ground truth for pixel-level reconstruction. This rigid replication and its associated image-level metrics~\citep{deng2026visrefiner, yang2025ui2code, li2025relook} fail in our setting because they force the code model to mimic the hallucinated content of the generated prior and penalize valid design variations. Our fundamental setting therefore challenges the code model to strictly satisfy the prompt's content while flexibly extracting layout and style from a noisy reference image.

Simply providing the reference image as an input is insufficient. Off-the-shelf vision-language models (VLMs) naturally entangle visual features with text semantics. When given a noisy reference, they tend to either reproduce its hallucinated content or ignore its layout entirely. The model therefore requires explicit training to extract the aesthetic prior while adhering to the text prompt. To facilitate this, we propose \method, which conceptualizes the generated target as a Design Graph. This structured intermediate representation explicitly separates semantic information from visual cues. Within this graph, nodes represent distinct functional elements such as text blocks, tables, and charts. Each node possesses decoupled attributes for its required content (e.g., specific strings or data series) and its visual properties (e.g., bounding boxes and colors). Edges explicitly define structural relationships, including hierarchical containment and spatial alignment.

Because the graph structure makes individual properties explicitly attributable, it naturally enables decoupled reward channels tailored to distinct objectives. Blending mixed outcomes into a single score (e.g., rendering correct text but a chaotic layout) often misaligns training gradients. To preserve specific learning signals, we derive synergic cross-modal rewards directly from the graph: deterministic verifiers provide fast, absolute checks for properties readable from code and DOM rendering, while a VLM judge assesses aesthetics through a \emph{Visual Graph Rubric} (VGR), avoiding unstable holistic scoring by answering node-specific binary questions. We then balance these channels using Group reward-Decoupled Normalization Policy Optimization (GDPO)~\citep{liu2026gdpo}, preventing dense verification signals from overwhelming sparse aesthetic feedback.

We also release an infographic benchmark evaluating verifiable correctness and visual quality separately, with automated visual scores closely matching human judgments. Ablations confirm that reference images consistently improve visual quality as an aesthetic prior across all tested models. Given the same references, our \method-8B and \method-32B gain 31.1 and 22.4 Visual points over their Qwen3-VL backbones, achieving visual quality comparable to frontier models like GPT-5.5 and Claude Opus~4.8 while making noticeably fewer content errors.

Our contributions are summarized as follows:
\begin{itemize}[leftmargin=2em,itemsep=2pt,topsep=2pt]
\item We introduce a novel paradigm for aesthetic code generation that utilizes image generators as an aesthetic prior. A reference image generated from the same prompt conditions the code model as input and provides visual signals during training.
\item We propose \method, which introduces decoupled cross-modal rewards driven by a design graph. It pairs deterministic verifiers for structural correctness with a sample-specific, graph-based VLM judge for aesthetic quality.
\item We release an infographic benchmark that evaluates correctness and visual quality separately, with human-aligned visual scores. References help every model we test, and \method matches leading proprietary models visually while following the prompt more closely.
\end{itemize}
\section{Related Work}
\label{sec:related_work}

\noindent\textbf{Visual Design and Code Generation.}
Posters, slides, and infographics can be generated as images or code. Image generators such as PosterMaker \citep{gao2025postermaker} and BizGen \citep{peng2025bizgen} compose appealing pages, but still misrender text and data on dense content \citep{tang2026igenbench}, and their raster outputs cannot be edited. Code is editable and verifiable. When written from text alone, as in LIDA \citep{dibia2023lida} and PPTAgent \citep{zheng2025pptagent}, it is composed without seeing the page and often looks plain. When written from an image, as in Design2Code \citep{si2025design2code} and ChartMimic \citep{yang2025chartmimic}, it copies the input as ground truth. Closest to ours, DesignAsCode \citep{liu2026designascode} and Editable Visual Design \citep{ye2026editable} let a coding agent consult a generated image, but only at inference time; the code model is never trained to use it. We instead train the code model to follow a prompt-generated image as an \emph{aesthetic prior} for layout and style.

\noindent\textbf{Reward Design for Visual Generation.}
RL training requires a reward for the rendered page. In image-to-code, this reward is usually similarity to the input image, measured with visual metrics \citep{rodriguez2025rlrf, chen2025rrvf} or MLLM critics \citep{yang2025ui2code, li2025relook}. For a generated reference, however, this would penalize valid alternatives and reward copied errors. Without a target image, rule-based rewards are precise but limited to what a program can check, such as element overlap in AeSlides \citep{pan2026aeslides}, while holistic MLLM judges \citep{zheng2023judging, chen2024mllm} give scores that shift with prompt phrasing. A more stable alternative breaks one judgment into many yes/no questions, first for text-to-image evaluation \citep{cho2024dsg} and more recently as rubric rewards for RL \citep{gunjal2025rubrics, ran2026ruler}. These questions, however, are derived from text alone and cannot check the design a reference proposes. We therefore build a \emph{design graph} from the prompt and the reference, verify code-level properties with rules, and turn the remaining visual properties into a \emph{Visual Graph Rubric} (VGR) tied to graph elements.

\noindent\textbf{Multi-Objective Reinforcement Learning.}
Critic-free methods such as GRPO \citep{shao2024deepseekmath} normalize rewards within a sampled group and, following DeepSeek-R1 \citep{guo2025deepseek}, have become a standard recipe for post-training LLMs and VLMs \citep{liu2025visual}. They assume a single reward, however: when several rewards are summed before normalization, a dense, near-saturated rule reward can drown out a sparse aesthetic one. MO-GRPO \citep{ichihara2025mogrpo} reweights rewards by their variance, while GDPO \citep{liu2026gdpo} normalizes each reward separately. We adopt GDPO to balance our rule-based and VGR rewards.

\Cref{sec:appendix_related_work} extends this review with a fuller discussion of each of the three threads above.


\section{Method}
\label{sec:method}

\method learns visual design through graph-structured supervision. We first construct a target design graph that combines prompt-required content with non-authoritative reference cues (\cref{sec:formulation}). We then route its properties to deterministic and perceptual reward channels (\cref{sec:distance}) and optimize them with SFT and GDPO~\citep{liu2026gdpo} (\cref{sec:training}). Finally, we define a separate evaluation protocol for factual correctness and human-aligned visual quality (\cref{sec:design_quality_evaluation}).

\subsection{Design Graph as a Supervision Target}
\label{sec:formulation}

The prompt $p$ specifies required information, while visual design admits many valid solutions. Learning needs a concrete signal, so an image generator supplies one compelling possibility $I_{\mathrm{ref}}$, which may be factually or structurally wrong. We therefore organize supervision as a design graph rather than a monolithic target, scoring prompt-derived requirements independently of the reference.

We define the supervision target as
\begin{equation}
  \mathcal G_{\mathrm{tgt}}=(\mathcal V,\mathcal E,\mathbf X_V,r)
  =\Phi(p,I_{\mathrm{ref}}),
\end{equation}
where the canvas root $r$ holds page-level attributes and contains every top-level element. The remaining nodes come from a closed vocabulary of atomic elements, such as text, images, charts, tables, and badges, and of containers, such as regions, cards, panels, lists, and groups, which hold other nodes instead of rendering content; three or more interchangeable siblings collapse into one group node with a count attribute. Attributes $\mathbf X_V$ record content, geometry, and appearance, including structured chart and table data, and each edge $(u,\tau,v)\in\mathcal E$ carries a relation type $\tau$ covering containment, alignment, order, adjacency, and semantic connection. This decomposition lets prompt requirements and reference cues constrain complementary properties independently, making supervision attributable across semantic dimensions, and since it assumes only elements and their two-dimensional structure, it applies broadly to structured visual artifacts.

\begin{figure}[t]
  \centering
  \includegraphics[width=\linewidth]{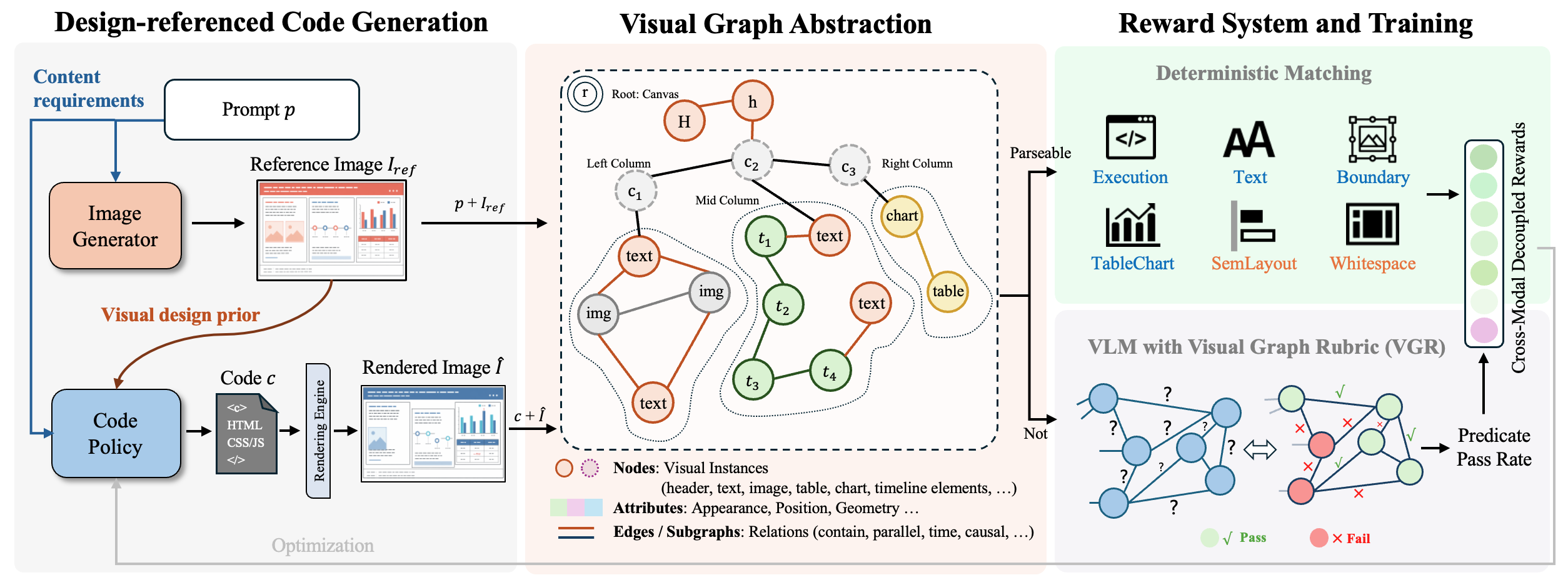}
    \caption{Overview of \method. Prompt requirements and reference design cues yield graph-based supervision. Conditioned on $p$ and $I_{\mathrm{ref}}$, the policy generates HTML/CSS code $c$, rendered as $\hat I$.}
  \label{fig:method_overview}
\end{figure}

From the generated code $c$ and its rendering $\hat I$, we jointly parse all candidate visual elements into $\hat{\mathcal G}=\Psi(c,\hat I)$. Code, DOM, and computed styles expose structured content and implementation details, while the rendering provides pixel geometry and visual evidence. We then compare $\hat{\mathcal G}$ with $\mathcal G_{\mathrm{tgt}}$ property by property to derive the reward signals below.

\subsection{From Design Graphs to Multi-Objective Rewards}
\label{sec:distance}

Graph properties divide by how they can be assessed. Those with structured evidence in code or measurable evidence in the rendering are programmatically verifiable, while appearance and higher-order relations vary by sample and require visual judgment. We therefore use deterministic verifiers and a sample-specific VLM-as-judge channel, detailed in \cref{sec:appendix_distance_illustration}.

\noindent\textbf{Deterministic verifiers.}
Six verifiers compute scores from code, the DOM, OCR, and rendered pixels. \emph{Execution} ($s_{\mathrm{exec}}$) checks renderability and fatal errors. \emph{Text} ($s_{\mathrm{text}}$) verifies required strings, and \emph{Boundary} ($s_{\mathrm{bdry}}$) measures canvas containment and aspect-ratio agreement. \emph{Table/Chart} ($s_{\mathrm{tc}}$) parses the structure, subtype, and data of HTML tables and ECharts charts, preventing static substitutes from receiving credit. \emph{Semantic Layout} ($s_{\mathrm{layout}}$) matches OCR text blocks between reference and candidate and scores their normalized positional agreement. \emph{Whitespace} ($s_{\mathrm{ws}}$) compares non-background occupancy at coarse and fine scales, capturing how whitespace is distributed.

\noindent\textbf{Visual Graph Rubric ($s_{\mathrm{vgr}}$).}
For the remaining properties, we derive a sample-specific list of Boolean questions $Q_{\mathrm{train}}$ from $\mathcal G_{\mathrm{tgt}}$. Each question binds to a node, attribute, relation, or subgraph, so the graph scaffolds rubric construction rather than serving as a policy input, turning open-ended visual judgment into explicit criteria. A VLM answers the fixed questions from the candidate rendering, and their pass rate gives
\begin{equation}
  s_{\mathrm{vgr}}=\tfrac{1}{|Q_{\mathrm{train}}|}\sum\nolimits_{q\in Q_{\mathrm{train}}}\mathbb{1}\!\left[\hat I\ \text{satisfies}\ q\right].
\end{equation}
Reference cues enter only through fixed questions, and VGR is used only for training.


\subsection{Training Pipeline}
\label{sec:training}

Cold-start SFT first establishes reliable prompt-and-image-to-code generation from filtered teacher demonstrations paired with the collected images. RL then switches to image-generator references and optimizes the seven rewards. These channels differ in applicability, scale, sparsity, saturation, and learning difficulty, so aggregating raw rewards can obscure useful signals. We therefore use Group reward-Decoupled Normalization Policy Optimization (GDPO)~\citep{liu2026gdpo}.

For each $(p,I_{\mathrm{ref}})$, we sample code rollouts $\mathcal B=\{c_i\}_{i=1}^{B}$, render $\hat I_i=\operatorname{render}(c_i)$, and parse $\hat{\mathcal G}_i=\Psi(c_i,\hat I_i)$. Comparing $\hat{\mathcal G}_i$ with $\mathcal G_{\mathrm{tgt}}$ produces the seven rewards above as $\mathbf s_i=(s_i^{(v)})_{v=1}^{7}\in[0,1]^7$.
A~mask $m_i^{(v)}\in\{0,1\}$ indicates whether channel $v$ applies to rollout $i$. Let $\mu_v$ and $\sigma_v$ denote its mean and standard deviation over the applicable subset $\mathcal B_v=\{i\mid m_i^{(v)}=1\}$. GDPO normalizes each channel before averaging the applicable advantages:
\begin{equation}
  A_i^{(v)}=\tfrac{s_i^{(v)}-\mu_v}{\sigma_v+\epsilon},
  \qquad
  A_i=\tfrac{\sum\nolimits_v m_i^{(v)}A_i^{(v)}}{\sum\nolimits_v m_i^{(v)}}.
  \label{eq:gdpo_advantage}
\end{equation}
With $\rho_i(\theta)=\pi_\theta(c_i\mid p,I_{\mathrm{ref}})/\pi_{\mathrm{old}}(c_i\mid p,I_{\mathrm{ref}})$, GDPO maximizes the clipped group objective
\begin{equation}
  \mathcal L_{\mathrm{GDPO}}(\theta)=\frac{1}{B}\sum\nolimits_{i=1}^{B}
  \min\!\left(\rho_i(\theta)A_i,\operatorname{clip}\bigl(\rho_i(\theta),1-\varepsilon,1+\varepsilon\bigr)A_i\right).
  \label{eq:gdpo_objective}
\end{equation}
Applicability masks exclude irrelevant checks, so all valid objectives contribute to one policy update without first collapsing their raw rewards into a scalar.

\definecolor{codeCol}{HTML}{1F4E79}
\definecolor{genCol}{HTML}{C55A11}
\newcommand{\na}{\textcolor{black!60}{\rule[0.42ex]{7pt}{0.6pt}}}

\definecolor{groupBasicCol}{HTML}{DDEBF7}
\definecolor{groupDesignCol}{HTML}{FCEAD9}
\definecolor{oursRow}{HTML}{F5F6F7}
\providecommand{\mainresultstablefont}{\fontsize{8.2}{9.8}\selectfont}
\providecommand{\mainresultstabcolsep}{2pt}
\providecommand{\mainresultscontentheader}{\textbf{Content}}

\newcommand{\mainresultsfloat}{
\begin{table}[t]
    \centering
    \caption{\textbf{Results on our infographic benchmark of 300 instances.} Models in the first two blocks receive only the prompt; those in the third, including \method, also receive a generated reference image. Rule averages the programmatic Text, Bound, and Chart checks; Visual averages the Content, Layout, and Style scores as defined in \cref{sec:design_quality_evaluation}. Overall is the mean of Rule and Visual. Scores (\%) are averaged over three generations per prompt; higher is better. Bold and underlined values indicate the best and second-best scores in each column.}
    \label{tab:model_performance}
    \mainresultstablefont
    \renewcommand{\arraystretch}{1.16}
    \setlength{\tabcolsep}{\mainresultstabcolsep}
    \begin{tabular}{@{}p{100pt}@{\hspace{4pt}}|@{\hspace{4pt}} *{3}{>{\centering\arraybackslash}p{28pt}} >{\columncolor{groupBasicCol}\centering\arraybackslash}p{28pt} *{3}{>{\centering\arraybackslash}p{28pt}} >{\columncolor{groupDesignCol}\centering\arraybackslash}p{28pt} >{\columncolor{oursRow}\centering\arraybackslash}p{28pt}@{}}
        \toprule
        \textbf{Model}
            & \textbf{Text} & \textbf{Bound} & \textbf{Chart} & \textbf{Rule}
            & \mainresultscontentheader & \textbf{Layout} & \textbf{Style} & \textbf{Visual} & \textbf{Overall} \\
        \midrule
        \footnotesize\textit{Frontier Closed-Source} & \multicolumn{9}{c}{} \\[-2pt]
        GPT-5.5              & 88.67 & 86.18 & 82.85 & 85.90 & 79.27 & 68.09 & 52.80 & \cellcolor{groupDesignCol}66.72 & 76.31 \\
        GPT-5.4              & 76.37 & 86.14 & 65.30 & 75.94 & 71.05 & 63.29 & 47.31 & \cellcolor{groupDesignCol}60.55 & 68.24 \\
        GPT-5.1              & 80.42 & 78.64 & 73.68 & 77.58 & 73.17 & 59.76 & 43.12 & \cellcolor{groupDesignCol}58.68 & 68.13 \\
        GPT-4o               & 73.64 & 78.37 & 55.64 & 69.22 & 51.34 & 33.33 & 15.95 & \cellcolor{groupDesignCol}33.54 & 51.38 \\
        Claude Opus 4.8      & 91.40 & 85.03 & 87.69 & 88.04 & 84.98 & 64.88 & 46.72 & \cellcolor{groupDesignCol}65.53 & 76.78 \\
        Gemini 3.1 Pro       & 69.95 & 74.47 & 67.61 & 70.68 & 74.68 & 60.73 & 41.00 & \cellcolor{groupDesignCol}58.81 & 64.74 \\
        \midrule
        \footnotesize\textit{Open-Source VLMs} & \multicolumn{9}{c}{} \\[-2pt]          Qwen3.8-27B           & 81.29 & 81.85 & 70.66 & 77.93 & 82.72 & 83.27 & 53.45 & \cellcolor{groupDesignCol}73.15 & 75.54 \\
        Qwen3.5-35B-A3B       & 52.08 & 72.95 & 41.56 & 55.53 & 70.57 & 71.41 & 43.27 & \cellcolor{groupDesignCol}61.75 & 58.64 \\
        Qwen3-VL-235B-A22B    & 68.94 & 79.69 & 54.54 & 67.73 & 56.43 & 39.49 & 22.48 & \cellcolor{groupDesignCol}39.46 & 53.59 \\
        Qwen3-VL-32B          & 66.00 & 80.31 & 41.36 & 62.55 & 49.81 & 34.85 & 19.34 & \cellcolor{groupDesignCol}34.67 & 48.61 \\
        Qwen3-VL-30B-A3B      & 65.20 & 77.24 & 50.27 & 64.24 & 46.44 & 33.10 & 19.35 & \cellcolor{groupDesignCol}32.96 & 48.60 \\
        Qwen3-VL-8B           & 58.32 & 77.23 & 43.30 & 59.61 & 38.33 & 23.97 & 13.22 & \cellcolor{groupDesignCol}25.17 & 42.39 \\
        Qwen2.5-VL-72B        & 74.24 & 83.63 & 43.48 & 67.12 & 39.34 & 22.37 & 9.00 & \cellcolor{groupDesignCol}23.57 & 45.34 \\
        InternVL3.5-241B-A28B & 82.01 & 69.44 & 67.19 & 72.88 & 69.95 & 65.12 & 33.96 & \cellcolor{groupDesignCol}56.34 & 64.61 \\
        InternVL3-78B         & 67.59 & 80.72 & 50.88 & 66.39 & 42.75 & 26.42 & 11.36 & \cellcolor{groupDesignCol}26.84 & 46.62 \\
        Kimi-VL               & 61.70 & 58.70 & 48.89 & 56.43 & 31.92 & 18.12 & 6.97 & \cellcolor{groupDesignCol}19.00 & 37.72 \\
        MiniCPM-V-4.5         & 61.04 & 72.03 & 38.81 & 57.29 & 46.05 & 40.98 & 21.34 & \cellcolor{groupDesignCol}36.12 & 46.71 \\
        Llama-4 Scout         & 76.59 & 81.49 & 66.93 & 75.00 & 58.51 & 54.82 & 26.47 & \cellcolor{groupDesignCol}46.60 & 60.80 \\
        \midrule
        \footnotesize\textit{With reference image input} & \multicolumn{9}{c}{} \\[-2pt]
        GPT-5.5               & 89.22 & 86.85 & 81.35 & 85.80 & 82.44 & \underline{89.93} & \textbf{57.91} & \cellcolor{groupDesignCol}76.76 & 81.28 \\
        Claude Opus 4.8       & 92.45 & 84.43 & 73.75 & 83.55 & \textbf{86.42} & 89.76 & 55.51 & \cellcolor{groupDesignCol}\underline{77.23} & 80.39 \\
        Qwen3.8-27B           & 82.77 & 83.31 & 71.96 & 79.34 & 85.07 & 87.59 & 54.49 & \cellcolor{groupDesignCol}75.72 & 77.53 \\
        Qwen3-VL-32B          & 71.19 & 79.61 & 50.77 & 67.19 & 62.04 & 64.59 & 38.43 & \cellcolor{groupDesignCol}55.02 & 61.10 \\
        Qwen3-VL-8B           & 64.07 & 70.99 & 42.35 & 59.14 & 52.90 & 51.33 & 29.92 & \cellcolor{groupDesignCol}44.72 & 51.93 \\
        Qwen2.5-VL-72B        & 73.02 & 81.98 & 41.81 & 65.60 & 53.39 & 43.22 & 18.95 & \cellcolor{groupDesignCol}38.52 & 52.06 \\
        InternVL3-78B         & 65.27 & 79.93 & 47.03 & 64.08 & 52.33 & 41.15 & 19.79 & \cellcolor{groupDesignCol}37.76 & 50.92 \\
        Kimi-VL               & 65.60 & 60.44 & 40.04 & 55.36 & 37.89 & 24.81 & 10.71 & \cellcolor{groupDesignCol}24.47 & 39.91 \\
        \addlinespace[1pt]
        \rowcolor{oursRow} \method-8B  & \underline{94.06} & \underline{88.36} & \underline{87.79} & \cellcolor{groupBasicCol}\underline{90.07} & \underline{86.41} & 87.80 & 53.21 & \cellcolor{groupDesignCol}75.80 & \cellcolor{oursRow}\underline{82.94} \\
        \rowcolor{oursRow} \method-32B & \textbf{95.34} & \textbf{97.27} & \textbf{90.37} & \cellcolor{groupBasicCol}\textbf{94.33} & 85.76 & \textbf{90.58} & \underline{55.99} & \cellcolor{groupDesignCol}\textbf{77.44} & \cellcolor{oursRow}\textbf{85.89} \\
        \bottomrule
    \end{tabular}
\end{table}
}
\providecommand{\floattoppad}{}
\providecommand{\figcappad}{}
\providecommand{\tabcappad}{}
\providecommand{\floatbotpad}{}

\newcommand{\qualitativecomparisonfloat}{
\begin{figure}[t]
    \centering
    \floattoppad
    \includegraphics[width=\linewidth]{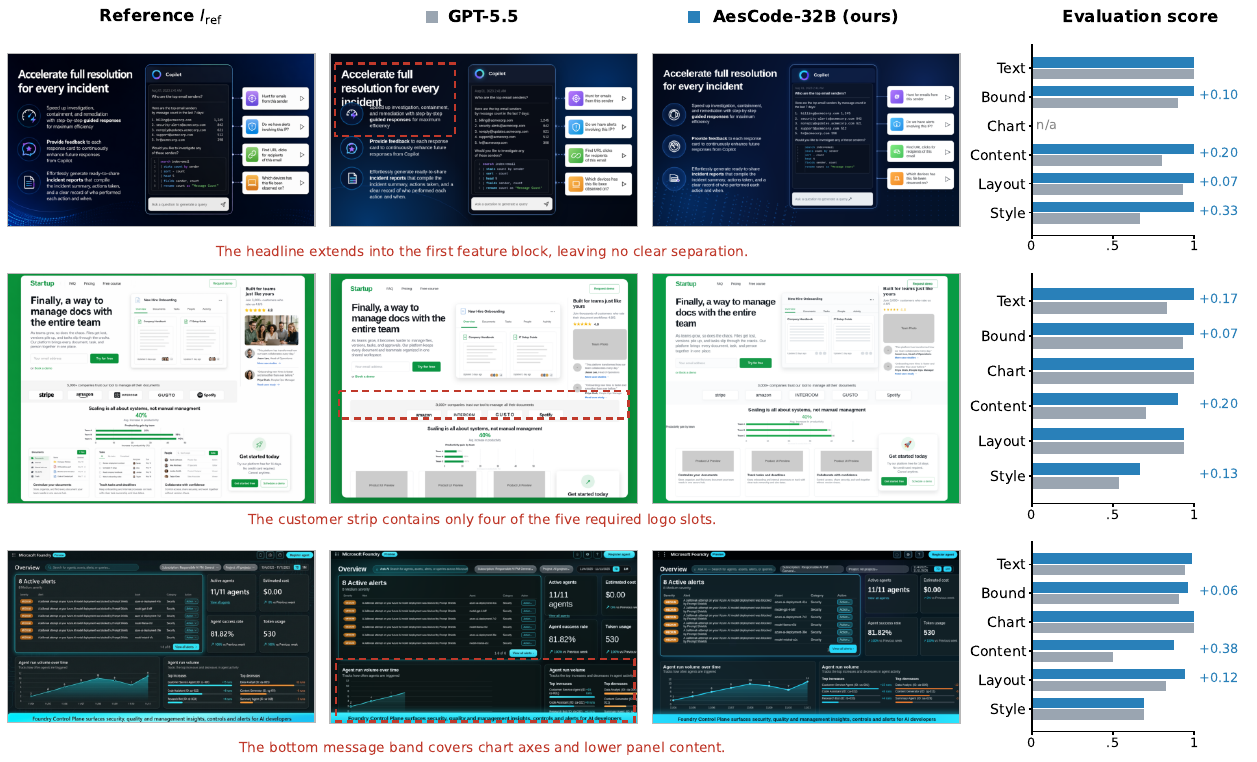}
    \figcappad
    \caption{\textbf{Qualitative comparisons selected by the three-rollout mean Overall gap.} Each row shows the reference and outputs from GPT-5.5 and \method-32B. Bars report evaluation scores; red dashed boxes mark visible GPT-5.5 errors.}
    \label{fig:qualitative_cases}
    \floatbotpad
\end{figure}
}

\newcommand{\externalbenchmarkfloat}{
\begin{table}[t]
\centering
\tabcappad
\caption{\textbf{Transfer to three external benchmarks.}
GDB/lica evaluates \textcolor{genCol}{text-to-image} generation; SlidesBench and
Design2Code evaluate \textcolor{codeCol}{image-to-code} reconstruction. All models
output rendered HTML/CSS under each benchmark's official protocol. SlidesBench uses
a 0--5 scale; Design2Code and GDB/lica CLIP, Harm., and OCR use 0--100. Higher is
better except FID.}
\label{tab:external_benchmarks}
\normalsize
\setlength{\tabcolsep}{2pt}
\renewcommand{\arraystretch}{1.15}
\resizebox{\linewidth}{!}{%
\begin{tabular}{@{}l *{6}{>{\centering\arraybackslash}p{28pt}} @{\hspace{6pt}} *{3}{>{\centering\arraybackslash}p{28pt}} @{\hspace{6pt}} *{5}{>{\centering\arraybackslash}p{28pt}}@{}}
\toprule
& \multicolumn{6}{c}{\textcolor{genCol}{\textbf{GDB/lica}}}
& \multicolumn{3}{c}{\textcolor{codeCol}{\textbf{SlidesBench}}}
& \multicolumn{5}{c}{\textcolor{codeCol}{\textbf{Design2Code}}} \\
\cmidrule(lr){2-7}\cmidrule(lr){8-10}\cmidrule(lr){11-15}
\textbf{Method} & NIMA & CLIP & Pick & FID$_{\downarrow}$ & Harm. & OCR
& Cont. & Layout & Color
& Block & Text & Pos. & Color & CLIP \\
\midrule
GPT-5.5           & \underline{4.362} & \textbf{30.2} & \textbf{19.53} & \underline{0.439} & 70.6 & \textbf{71.9}
                  & \textbf{4.54} & \textbf{2.85} & \underline{3.63}
                  & \textbf{92.3} & \textbf{98.5} & \textbf{94.2} & \textbf{95.3} & \textbf{94.3} \\
GPT-4o            & 2.589 & 24.1 & 17.82 & 0.673 & 69.3 & 43.3
                  & 3.41 & 2.00 & 2.81
                  & 77.5 & 93.0 & 77.9 & 82.1 & 89.5 \\
GLM-4.5V          & 3.682 & 26.4 & 18.49 & 0.541 & 70.3 & 65.1
                  & \underline{4.22} & 2.14 & 3.00
                  & 76.7 & 88.2 & 73.8 & 84.7 & 90.3 \\
Qwen2.5-VL-72B    & 3.564 & 26.0 & 18.40 & 0.537 & 69.1 & 62.6
                  & 4.03 & 2.02 & 2.83
                  & 82.6 & 97.3 & 78.0 & 78.8 & 88.9 \\
Qwen3-VL-30B-A3B  & 3.461 & 26.1 & 18.38 & 0.532 & 68.8 & 57.6
                  & 3.95 & 2.10 & 2.97
                  & 86.1 & 95.6 & 80.5 & 91.3 & 91.4 \\
Qwen3-VL-32B      & 3.122 & 25.4 & 18.29 & 0.522 & \underline{70.8} & 56.1
                  & 3.80 & 2.12 & 3.02
                  & 88.9 & 97.8 & 83.2 & 92.8 & 91.7 \\
Qwen3-VL-8B       & 2.821 & 23.7 & 17.85 & 0.647 & 68.7 & 43.7
                  & 3.64 & 1.97 & 2.81
                  & 74.0 & 86.8 & 70.6 & 79.3 & 88.7 \\
\addlinespace[1pt]
\rowcolor{oursRow} \method-32B
                  & \textbf{4.932} & \underline{29.8} & \underline{19.23} & \textbf{0.394} & \textbf{71.3} & \underline{68.5}
                  & 3.95 & \underline{2.49} & \textbf{3.92}
                  & \underline{90.0} & \underline{98.1} & \underline{86.4} & \underline{93.9} & \underline{92.9} \\
\bottomrule
\end{tabular}%
}
\floatbotpad
\end{table}
}

\newcommand{\ablationresultsfloat}{
\begin{table}[t]
    \centering
    \caption{\textbf{Reference images during training and inference.} All rows omit reference input at inference; the lower row also omits it from policy inputs during training while retaining reference-based rewards. Gray values show changes from reference-conditioned results.}
    \label{tab:noncond}
    \renewcommand{\arraystretch}{1.2}
    \setlength{\tabcolsep}{5pt}
    \resizebox{\linewidth}{!}{%
    \begin{tabular}{l@{\hspace{0.8em}} c c c >{\columncolor{groupBasicCol}}c c c c >{\columncolor{groupDesignCol}}c >{\columncolor{oursRow}}c}
        \toprule
        \textbf{Model}
            & \textbf{Text} & \textbf{Bound} & \textbf{Chart} & \textbf{Rule}
            & \textbf{Content} & \textbf{Layout} & \textbf{Style} & \textbf{Visual} & \textbf{Overall} \\
        \midrule
        \multicolumn{10}{l}{\small\textit{Reference input removed at inference only}} \\
        GPT-5.5      & 88.67\,{\scriptsize\color{gray}$(-0.6)$} & 86.18\,{\scriptsize\color{gray}$(-0.7)$} & \underline{82.85}\,{\scriptsize\color{gray}$(+1.5)$} & 85.90\,{\scriptsize\color{gray}$(+0.1)$} & 79.27\,{\scriptsize\color{gray}$(-3.2)$} & 68.09\,{\scriptsize\color{gray}$(-21.8)$} & \textbf{52.80}\,{\scriptsize\color{gray}$(-5.1)$} & \cellcolor{groupDesignCol}66.72\,{\scriptsize\color{gray}$(-10.0)$} & \cellcolor{oursRow}76.31\,{\scriptsize\color{gray}$(-5.0)$} \\
        Qwen3-VL-32B & 66.00\,{\scriptsize\color{gray}$(-5.2)$} & 80.31\,{\scriptsize\color{gray}$(+0.7)$} & 41.36\,{\scriptsize\color{gray}$(-9.4)$} & 62.55\,{\scriptsize\color{gray}$(-4.6)$} & 49.81\,{\scriptsize\color{gray}$(-12.2)$} & 34.85\,{\scriptsize\color{gray}$(-29.7)$} & 19.34\,{\scriptsize\color{gray}$(-19.1)$} & \cellcolor{groupDesignCol}34.67\,{\scriptsize\color{gray}$(-20.4)$} & \cellcolor{oursRow}48.61\,{\scriptsize\color{gray}$(-12.5)$} \\
        Qwen3-VL-8B  & 58.32\,{\scriptsize\color{gray}$(-5.8)$} & 77.23\,{\scriptsize\color{gray}$(+6.2)$} & 43.30\,{\scriptsize\color{gray}$(+1.0)$} & 59.61\,{\scriptsize\color{gray}$(+0.5)$} & 38.33\,{\scriptsize\color{gray}$(-14.6)$} & 23.97\,{\scriptsize\color{gray}$(-27.4)$} & 13.22\,{\scriptsize\color{gray}$(-16.7)$} & \cellcolor{groupDesignCol}25.17\,{\scriptsize\color{gray}$(-19.6)$} & \cellcolor{oursRow}42.39\,{\scriptsize\color{gray}$(-9.5)$} \\
        \rowcolor{oursRow} \method-8B & \underline{93.74}\,{\scriptsize\color{gray}$(-0.3)$} & \underline{88.19}\,{\scriptsize\color{gray}$(-0.2)$} & 82.36\,{\scriptsize\color{gray}$(-5.4)$} & \cellcolor{groupBasicCol}\underline{88.09}\,{\scriptsize\color{gray}$(-2.0)$} & \textbf{84.92}\,{\scriptsize\color{gray}$(-1.5)$} & \textbf{86.84}\,{\scriptsize\color{gray}$(-1.0)$} & \underline{52.63}\,{\scriptsize\color{gray}$(-0.6)$} & \cellcolor{groupDesignCol}\textbf{74.80}\,{\scriptsize\color{gray}$(-1.0)$} & \cellcolor{oursRow}\textbf{81.45}\,{\scriptsize\color{gray}$(-1.5)$} \\
        \midrule
        \multicolumn{10}{l}{\small\textit{Reference input removed during training and inference}} \\
        \rowcolor{oursRow} \method-8B & \textbf{94.04} & \textbf{88.55} & \textbf{84.43} & \cellcolor{groupBasicCol}\textbf{89.01} & \underline{81.39} & \underline{80.04} & 47.71 & \cellcolor{groupDesignCol}\underline{69.71} & \cellcolor{oursRow}\underline{79.36} \\
        \bottomrule
    \end{tabular}
    }
    \vspace{0.76\baselineskip}
    \centering
    \caption{\textbf{Method ablations on Qwen3-VL-8B with reference input.} $\checkmark$ and $\times$ denote included and removed components; named entries denote replacements. Holistic VLM score is a single absolute rating, direct rubric authors the visual questions directly from the inputs without first constructing the design graph, and Layout/WS indicates the Semantic Layout and Whitespace rewards.}
    \label{tab:ablation_main}
    \renewcommand{\arraystretch}{1.1}
    \setlength{\tabcolsep}{5pt}
    \resizebox{0.78\linewidth}{!}{%
    \begin{tabular}{l c c c c c >{\columncolor{groupBasicCol}}c >{\columncolor{groupDesignCol}}c >{\columncolor{oursRow}}c}
        \toprule
        \textbf{Component} & \textbf{SFT} & \textbf{RL} & \textbf{GDPO} & \textbf{Layout/WS} & \textbf{VGR} & \textbf{Rule} & \textbf{Visual} & \textbf{Overall} \\
        \midrule
        \rowcolor{oursRow} \textbf{\method (full)} & $\checkmark$ & $\checkmark$ & $\checkmark$ & $\checkmark$ & $\checkmark$ & \cellcolor{groupBasicCol}90.07 & \cellcolor{groupDesignCol}75.80 & \cellcolor{oursRow}82.94 \\
        \midrule
        \multirow{2}{*}{Training pipeline}
            & $\checkmark$ & $\times$ & -- & -- & -- & 79.28 & 65.86 & 72.57 \\
            & $\times$ & $\checkmark$ & $\checkmark$ & $\checkmark$ & $\checkmark$ & 86.09 & 69.85 & 77.97 \\
        \midrule
        Reward aggregation
            & $\checkmark$ & $\checkmark$ & GRPO & $\checkmark$ & $\checkmark$ & 90.32 & 70.68 & 80.50 \\
        \midrule
        \multirow{5}{*}{Design supervision}
            & $\checkmark$ & $\checkmark$ & $\checkmark$ & $\times$ & $\times$ & 90.47 & 66.59 & 78.53 \\
            & $\checkmark$ & $\checkmark$ & $\checkmark$ & $\checkmark$ & $\times$ & 91.03 & 68.12 & 79.58 \\
            & $\checkmark$ & $\checkmark$ & $\checkmark$ & $\checkmark$ & CLIP & 89.14 & 70.49 & 79.82 \\
            & $\checkmark$ & $\checkmark$ & $\checkmark$ & $\checkmark$ & Holistic VLM score & 89.20 & 68.91 & 79.06 \\
            & $\checkmark$ & $\checkmark$ & $\checkmark$ & $\checkmark$ & Direct rubric & 90.05 & 72.61 & 81.33 \\
        \bottomrule
    \end{tabular}
    }
\end{table}
}

\newcommand{\rewarddynamicsfloat}{
\begin{figure}[t]
  \centering
  \includegraphics[width=\linewidth]{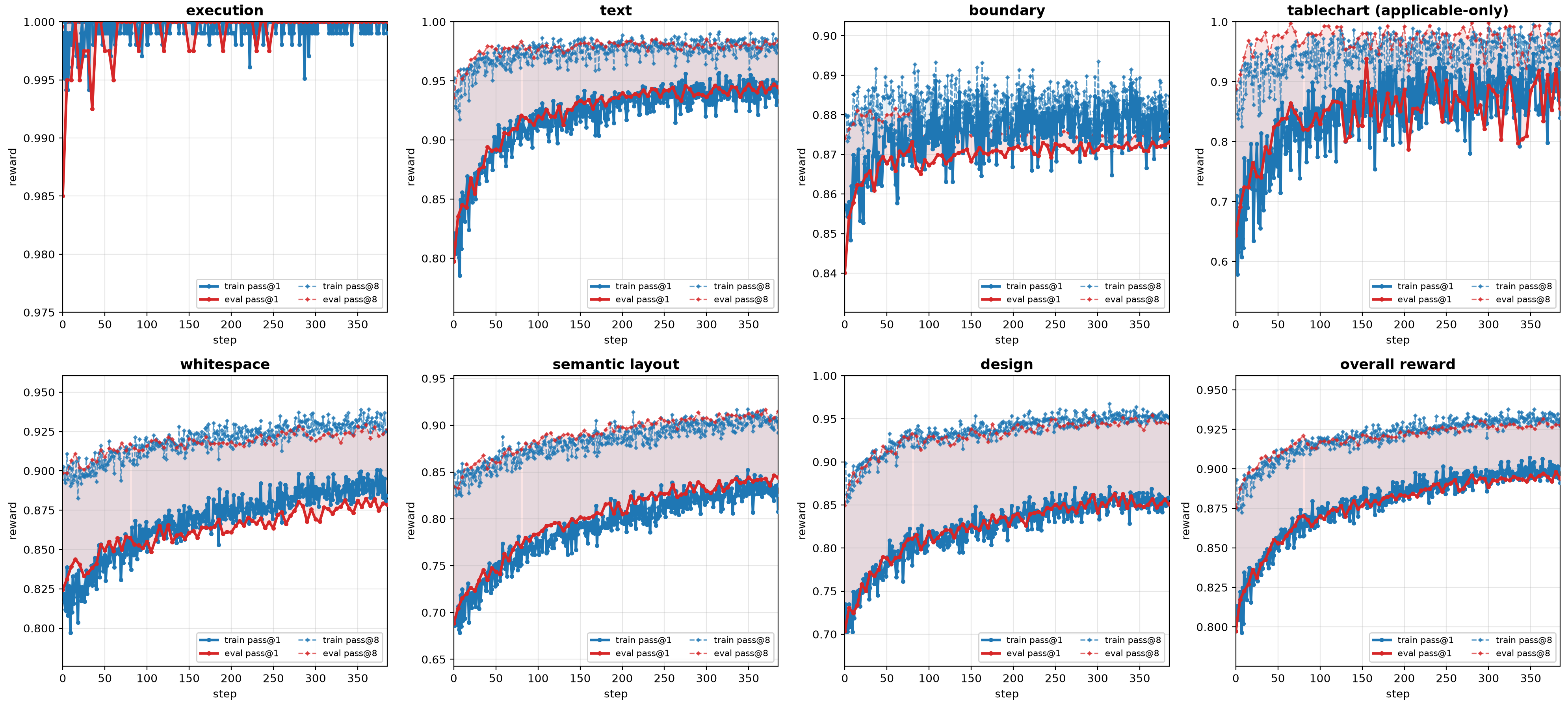}
    \caption{\textbf{Per-channel training rewards under GDPO.} Solid: single rollout; dashed: mean over eight rollouts; blue: training set; red: held-out set.}
  \label{fig:post_training_dynamics}
\end{figure}
}

\newcommand{\humanstudyfloat}{
\ifdefined\microsoftanalysisfloats
\else
\begin{figure}[t]
  \centering
   \includegraphics[width=\linewidth]{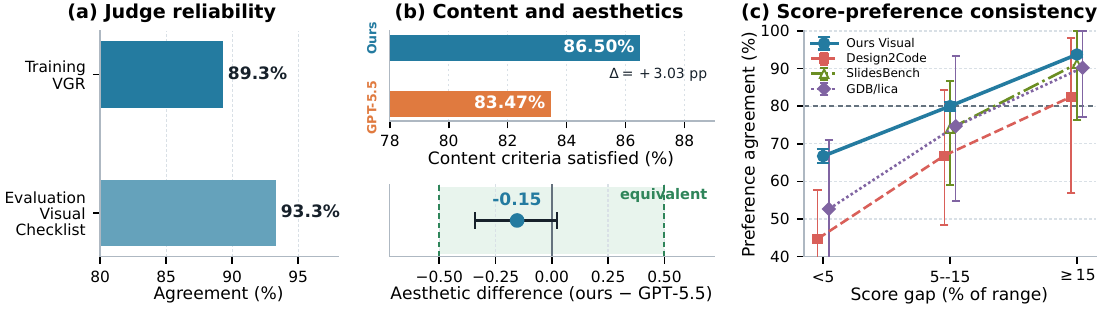}
        \caption{\textbf{Human evaluation.} \emph{(a)} Agreement with training VGR and evaluation-checklist judgments. \emph{(b)} Content satisfaction and aesthetic equivalence between \method-32B and GPT-5.5. \emph{(c)} Agreement between metric rankings and human preference by normalized score gap.}
  \label{fig:user_study}
\end{figure}
\fi
}

\newcommand{\failurecasesfloat}{
\ifdefined\microsoftanalysisfloats
\begin{figure}[t]
    \centering
    \includegraphics[width=\linewidth]{figures/user_study_150_final.pdf}
    \caption{\textbf{Human evaluation.} \emph{(a)} Agreement with training VGR and evaluation-checklist judgments. \emph{(b)} Content satisfaction and aesthetic equivalence between \method-32B and GPT-5.5. \emph{(c)} Agreement between metric rankings and human preference by normalized score gap.}
    \label{fig:user_study}
    \vspace{1.2em}
    \includegraphics[width=\linewidth]{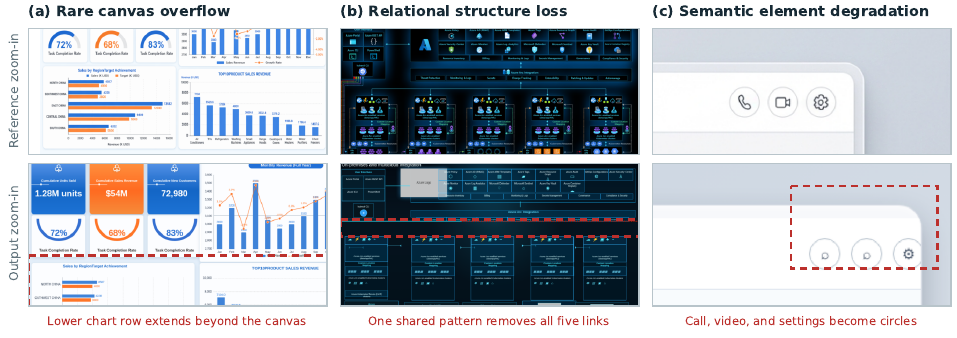}
    \caption{\textbf{Residual failures of \method-32B.} Reference and output zoom-ins show \emph{(a)} canvas overflow, \emph{(b)} lost relational structure, and \emph{(c)} semantic element degradation. Dashed boxes mark affected regions; frequencies and additional cases appear in \cref{appendix:failure_analysis}.}
    \label{fig:failure_cases}
\end{figure}
\else
\begin{figure}[t]
  \centering
        \includegraphics[width=\linewidth]{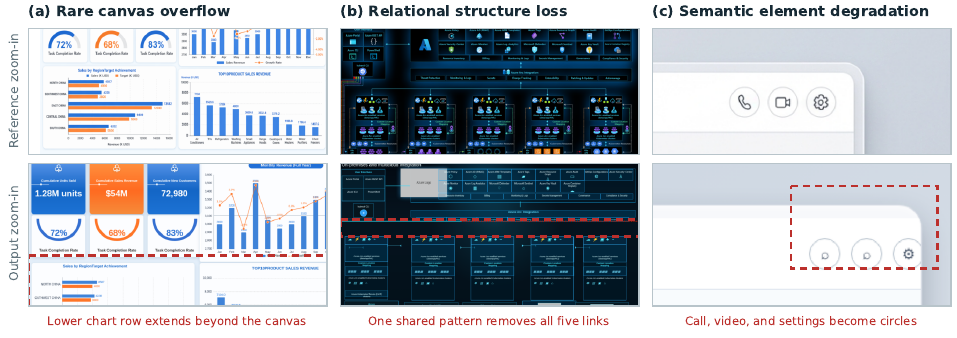}
                                \caption{\textbf{Residual failures of \method-32B.} Reference and output zoom-ins show \emph{(a)} canvas overflow, \emph{(b)} lost relational structure, and \emph{(c)} semantic element degradation. Dashed boxes mark affected regions; frequencies and additional cases appear in \cref{appendix:failure_analysis}.}
  \label{fig:failure_cases}
\end{figure}
\fi
}

\subsection{Final Evaluation Protocol}
\label{sec:design_quality_evaluation}

Training requires a concrete reference, but evaluation must accept multiple valid designs. We therefore assess the artifact itself through \textbf{Rule}, \textbf{Visual}, and their mean \textbf{Overall}.

\noindent\textbf{Rule.}
Rule averages Text, Boundary, and Table/Chart (\cref{sec:distance}), which verify exact content, canvas containment, and structured data from code and rendering.

\noindent\textbf{Visual.}
Visual converts open-ended, sample-specific quality into a checklist $Q_p$ derived from prompt requirements. \emph{Content} checks required-element presence and completeness; \emph{Layout} covers organization, relations, and collisions; and \emph{Style} judges hierarchy, typography, color, and component finish against a production-ready standard. Each checklist averages 35.8 questions. A VLM answers fixed yes/no questions from the candidate rendering, and Visual averages the three dimensions. These are absolute criteria for the artifact rather than pairwise comparisons. We develop the pipeline on a separate 50-instance set: it drafts questions, filters ambiguous, non-visual, and implementation-specific criteria, then revises wording and coverage. Annotators review questions and rank output pairs; ranking disagreements guide each revision. Once consistency exceeds $90\%$ and no issues remain, we freeze the pipeline and apply it unchanged to the disjoint 300-instance benchmark, giving every system the same sample-specific criteria. \Cref{appendix:evaluation_protocol} details the pipeline, while \cref{appendix:human_study} confirms strong agreement with human aesthetic judgments and pairwise choices.

\section{Experiments}
\label{sec:experiments}

\subsection{Experimental Setup}
\label{sec:setup}

\noindent\textbf{Training data.}
We curate a candidate pool of approximately 18{,}000 human-designed slides, posters, dashboards, and webpages, clustered by their CLIP image features. For each sampled image, GPT-5.4-mini infers a generation prompt that specifies required contents (e.g., text and charts) without prescribing visual layout or styles. For SFT, we construct 3{,}000 demonstrations using GPT-5.5 paired with the original human designs. We generate three rollouts per prompt--image pair and keep the highest-scoring candidate if its reward exceeds $0.85$, otherwise resampling from the same cluster. Retaining original designs for SFT teaches the prompt-and-reference-to-code format without image-generation overhead. For RL, we score eight rollouts from the cold-start policy on every remaining prompt, keep the prompts whose $\text{pass}@1$ and $\text{pass}@8$ rates leave usable learning signal, rebalance them across clusters, and replace the original images with synthetic references from GPT-Image-2, yielding roughly 8{,}000 prompts. See details in \cref{appendix:data_construction}.

\noindent\textbf{Benchmark and baselines.}
Our test set for benchmarking contains 300 prompts, with GPT-Image-2 references (\cref{appendix:benchmark}). We compare proprietary and open-source models under prompt-only and reference-conditioned settings. Final evaluation follows the protocol in \cref{sec:design_quality_evaluation} (detailed checklist construction and a complete example checklist appear in \cref{appendix:visual_checklist}). We additionally evaluate Design2Code~\citep{si2025design2code}, SlidesBench~\citep{ge2025autopresent}, and GDB/lica~\citep{deganutti2026} using their respective protocols (\cref{appendix:external}).

\noindent\textbf{Implementation details.}
\method uses Qwen3-VL-8B and Qwen3-VL-32B, with SFT at a learning rate of $10^{-5}$ followed by GDPO using AdamW at $5\times10^{-6}$. Each RL step samples $128$ prompts with $8$ rollouts each; the 8B and 32B models train for $400$ and $520$ steps, respectively. Remaining optimization, decoding, and rendering settings appear in \cref{appendix:implementation}.

\mainresultsfloat

\subsection{Main Results}
\label{sec:main_results}

\noindent\textbf{Main observations.}
\Cref{tab:model_performance} reveals three main observations.
First, reference images improve visual presentation for all closed and open models by providing explicit layout and aesthetic anchors. Yet, Style remains a shared bottleneck below $60\%$: this metric gives credit only when a design requires no further visual revision before production delivery.
Second, models leverage visual guidance differently. Claude Opus 4.8 excels in content fidelity and uses references to elevate layout and style, whereas GPT-5.5 starts with stronger style and benefits most in content and layout. Most open models are highly imbalanced: some satisfy code-level constraints but remain weak visually, while others struggle on both. Qwen3.8 and Qwen3.5 stand out for more balanced visual planning, with Qwen3.8 showing the best balance between Rule and Visual among the open models evaluated. Its relatively high prompt-only Visual score leaves limited room for further gains from an additional reference ($73.15\to75.72$).
Third, our training substantially improves both base backbones, raising Overall from $51.93$ to $82.94$ at 8B and from $61.10$ to $85.89$ at 32B. The 32B model also leads Table/Chart at $90.37$. We require tables to use HTML table structures and charts to use ECharts, which constrains the solution space to executable, directly verifiable implementations and provides clear rewards for optimization. Overall, \method leads in verifiable correctness while matching frontier models such as GPT-5.5 and Claude Opus 4.8 in visual quality.

\qualitativecomparisonfloat
\externalbenchmarkfloat

\paragraph{Case studies.}
\Cref{fig:qualitative_cases} illustrates why frontier models can still trail \method despite plausible overall compositions. GPT-5.5 overlooks a subtle headline overlap, renders four rather than five requested logo slots, and lets a message band obscure a correctly instantiated ECharts chart. The last case retains a full Chart score because its structure and data are present; the failure lies in rendered visibility and is captured by Content and Layout. Together, these cases expose weaknesses in overlap control, requirement completeness, and the visual realization of valid code.

\ablationresultsfloat

\paragraph{Transfer to external benchmarks.}
Across three benchmarks with different tasks and metrics, \method-32B improves on its Qwen3-VL-32B backbone in all fourteen measures, demonstrating transfer beyond our benchmark. The largest gains appear on design-sensitive measures, including SlidesBench layout and color and GDB/lica NIMA, FID, and color harmony. Gains are smaller on Design2Code, an older website-reconstruction benchmark whose similarity metrics are near saturation and align less closely with human aesthetic judgments (\cref{fig:user_study}). Thus, GPT-5.5 remains stronger in exact screenshot fidelity, while \method shows that its transferable gains lie in layout and visual design rather than reconstruction-specific similarity.

\subsection{Ablation Study}
\label{sec:ablation}

\paragraph{Reference conditioning during training and inference.}
\Cref{tab:noncond} shows that reference-conditioned training internalizes multimodal visual planning into the policy. Removing the reference only at inference reduces \method-8B Visual by $1.00$ point, compared with $19.55$ for its backbone and $10.04$ for GPT-5.5. Even when the policy never sees the reference, reference-based rewards raise prompt-only Visual from $25.17$ to $69.71$, showing that such supervision transfers visual capability indirectly. This remains $5.09$ points below reference-conditioned training under the same inference setting, confirming the value of the reference as an input condition.

\paragraph{Training and reward design.}
\Cref{tab:ablation_main} shows that RL drives most gains, while SFT gives a stronger initialization: under the same RL budget, Overall rises from $77.97$ to $82.94$. GDPO adds $5.12$ Visual points over scalar GRPO by normalizing channels independently, which helps when verifiable and perceptual objectives learn at different rates (\cref{sec:analysis_rft_dynamics}). Design supervision then determines visual quality: code correctness alone yields $66.59$ Visual, and Layout/Whitespace raises it to $68.12$. CLIP reaches $70.49$ through coarse image-level guidance but cannot localize failed properties, while a holistic VLM score reaches only $68.91$ because its absolute signal is sparse and unstable. VGR decomposes visual quality into explicit graph-bound criteria and reaches $75.80$ without sacrificing Rule. Graph construction accounts for part of this: with the same inputs and training setup, authoring the questions directly, without the graph, lowers Visual to $72.61$ at unchanged Rule ($90.07$ vs.\ $90.05$).

\humanstudyfloat
\failurecasesfloat

\subsection{Analysis}
\label{sec:analysis}

\noindent\textbf{Human evaluation.}
\label{sec:analysis_user_study}%
Our blinded human study validates the automatic judges and the main comparison (\cref{appendix:human_study}). In \cref{fig:user_study}a, human-majority judgments agree with the training VGR and the Visual checklist on 89.3\% and 93.3\% of sampled predicates. In \cref{fig:user_study}b, \method-32B achieves higher Content satisfaction than GPT-5.5 (86.50\% vs.\ 83.47\%) with equivalent aesthetic quality (mean difference $-0.15$, 95\% CI $[-0.34,0.03]$ within the pre-specified $\pm0.5$ margin). In \cref{fig:user_study}c, our Visual score agrees with direct human pairwise choices in 80.66\% of comparisons, exceeds external metrics across comparable gap bins, and improves with larger score gaps.

\noindent\textbf{Failure analysis.}
\label{sec:analysis_failure}%
Lower-performing samples show three recurring failure modes (\cref{fig:failure_cases}). First, canvas overflow persists, though it is rare after training: code generation cannot reliably anticipate final pixel geometry, leaving visually grounded refinement to future work. Second, relational structure is harder than individual elements, since the model must understand content relations and render them as spatial connections, yet relation-level rewards stay sparse and do not localize the failed edge. Third, semantic elements often degrade into generic marks, since such icons require fine-grained SVG drawing grounded in meaning.

\section{Conclusion}\label{sec:conclusion}
This work asks how code models can generate information-rich visual artifacts that are both verifiable and visually compelling. We address this challenge by representing each target as a design graph that captures element content, visual attributes, and relations. Prompt requirements define what must be conveyed, while image-generated references guide how it should look; together, they yield targeted rewards that help compact code models internalize cross-modal understanding, organization, and visual judgment. \method matches frontier systems in human-aligned visual quality and surpasses them in verifiable correctness, with much of this capability retained even without reference input at inference. More broadly, our results show that image generators can serve not only as content generators, but also as scalable teachers of visual design for structured generative models. Remaining challenges in pixel-level geometry, complex relations, and semantic visual elements call for visually grounded iterative refinement and more precise structural feedback.

\FloatBarrier  
\bibliography{reference}
\bibliographystyle{unsrtnat}

\makeatletter
\let\ms@xfloat\@xfloat
\def\@xfloat#1[#2]{\ms@xfloat#1[#2htbp]}
\makeatother
\renewcommand{\topfraction}{0.95}
\renewcommand{\bottomfraction}{0.9}
\renewcommand{\textfraction}{0.05}
\renewcommand{\floatpagefraction}{0.85}
\setcounter{topnumber}{3}
\setcounter{bottomnumber}{2}
\setcounter{totalnumber}{5}

\makeatletter
\let\ms@clearpage\clearpage
\newcommand{\ms@softbarrier}{%
  \par
  \begingroup
    \let\clearpage\ms@clearpage
    \FloatBarrier
  \endgroup
}
\renewcommand{\clearpage}{\ms@softbarrier}
\makeatother

\fontsize{10pt}{12.6pt}\selectfont
\setlength{\parskip}{4pt}
\setlength{\textfloatsep}{10pt plus 2pt minus 3pt}
\setlength{\floatsep}{8pt plus 2pt minus 2pt}
\setlength{\intextsep}{8pt plus 2pt minus 2pt}
\captionsetup{skip=5pt}
\titlespacing*{\section}{0pt}{1.0em}{0.45em}
\titlespacing*{\subsection}{0pt}{0.75em}{0.3em}
\titlespacing*{\paragraph}{0pt}{0.5em}{0.8em}
\newcommand{\vgrexamplewidth}{0.74\linewidth}
\newcommand{\vgrportraitwidth}{0.66\linewidth}
\newcommand{\evalexamplewidth}{0.74\linewidth}
\newcommand{\appendixtocbreak}{\newpage}

%
%
\newpage
\appendix
\setcounter{table}{0}
\setcounter{figure}{0}
\setcounter{section}{0}
\numberwithin{table}{section}
\numberwithin{figure}{section}
\numberwithin{equation}{section}
\renewcommand{\thetable}{\Alph{section}\arabic{table}}
\renewcommand{\thefigure}{\Alph{section}\arabic{figure}}
\renewcommand{\theequation}{\Alph{section}\arabic{equation}}
\renewcommand{\thesection}{\Alph{section}}

\ifappendixtoc
  \addtocontents{toc}{\protect\setcounter{tocdepth}{2}}
  \begingroup
    \setlength{\parskip}{0pt}
    \hypersetup{linkcolor=black}  
    \renewcommand{\contentsname}{Appendix Contents}
    \tableofcontents
  \endgroup
  \providecommand{\appendixtocbreak}{\newpage}%
  \appendixtocbreak
\fi

\section{Extended Related Work}
\label{sec:appendix_related_work}

\noindent\textbf{Visual design and code generation.}
Image generators now produce product posters \citep{gao2025postermaker} and article-level infographics \citep{peng2025bizgen}, and COLE \citep{jia2023cole} assembles multi-layered designs from a pipeline of specialized models rather than general-purpose code. On dense infographics, however, they still distort data encodings and misspell strings \citep{tang2026igenbench}, and even a correct output is a raster that cannot be edited or checked element by element. Generating code avoids this. Text-to-code methods write visualization code \citep{dibia2023lida}, layouts \citep{feng2023layoutgpt}, or slides \citep{ge2025autopresent, zheng2025pptagent}, but compose the page without seeing it. Image-to-code methods recover the code behind a rendered webpage \citep{beltramelli2018pix2code, laurenccon2024unlocking, si2025design2code}, chart \citep{wu2025plot2code, yang2025chartmimic}, or SVG \citep{rodriguez2025starvector}. Because the input is the rendering of an existing program, a closer copy is always a better answer.

The closest work combines an image model with code at inference time. PosterVerse renders HTML typography over a generated background \citep{liu2026posterverse}, DesignAsCode uses an image-editing model to propose an improved render during reflection \citep{liu2026designascode}, and, concurrently, Editable Visual Design lets a coding agent use a generated visual as a global reference for composition and color \citep{ye2026editable}. In each case, the image steers a prompted pipeline, and the code model is never trained to use it. \method instead trains the code model to follow a prompt-generated reference as an aesthetic prior while ignoring its content errors, and turns the same reference into a training signal.

\noindent\textbf{Reward design for visual generation.}
Image-to-code RL rewards a rollout by how closely its render matches the input image, using visual similarity \citep{chen2025rrvf, rodriguez2025rlrf} or MLLM critics \citep{deng2026visrefiner, yang2025ui2code, li2025relook}. This is valid only when the input is an exact target. A generated reference is not: many programs realize the same prompt, and the generator may alter requested content. Without a target image, rule-based rewards such as layout metrics for slides \citep{pan2026aeslides} are stable but cover only what a program can decide. Preference-trained reward models \citep{xu2023imagereward, kirstain2023pick} capture subjective quality but inherit labeler noise. Holistic MLLM judges \citep{zheng2023judging, chen2024mllm} need no training, but their scores vary with prompt phrasing, and when used as an aesthetic reward, as in WebGen-R1 \citep{jiang2026webgenr1}, they reflect general quality rather than what suits a particular prompt.

Judgments become more reliable when split into narrow yes/no questions. TIFA \citep{hu2023tifa} and DSG \citep{cho2024dsg} do so for text-to-image evaluation, ArtifactsBench uses per-task checklists to evaluate rendered artifacts \citep{zhang2025artifactsbench}, and instruction-derived rubrics now serve as RL rewards for language models \citep{viswanathan2025checklists, gunjal2025rubrics} and, concurrently, for SVG \citep{ran2026ruler}. All of these derive their questions from text alone. Our Visual Graph Rubric derives them from a design graph of both the prompt and the reference, so it can check the layout and style the reference proposes, and each failure points to a specific element. Properties a program can decide are left to rule-based verifiers.

\noindent\textbf{Multi-objective reinforcement learning.}
Post-training RL has moved from PPO \citep{schulman2017proximal, ouyang2022training} to critic-free, group-normalized methods such as GRPO \citep{shao2024deepseekmath}, and we follow the R1-style recipe of a supervised cold start followed by such RL \citep{guo2025deepseek, liu2025visual}. With seven reward channels, summing them before group normalization lets their different scales and firing rates interfere, so a dense, near-saturated channel can suppress a sparse but informative one. MO-GRPO reweights rewards by their variance \citep{ichihara2025mogrpo}, while GDPO normalizes each channel separately \citep{liu2026gdpo}. We adopt GDPO because our channels differ widely in how often they fire and how much they vary.

\section{Data and Benchmark Construction}
\label{sec:appendix_dataset_construction}

\subsection{Source Designs and Stratified Sampling}
\label{appendix:data_construction}

Every split is built from human-designed graphics, because content is what they supply best: the text has a reason to be there, the sections stand in a sensible relation to one another, and the reading order reflects an actual communicative intent. Prompts and verifiable requirements are therefore inferred from human designs for the SFT, RL, and benchmark splits alike. Visual style is a separate matter. Designs that are modern and polished in every respect are a small and unevenly distributed minority of the collected pool, so the style of a given source image is largely an accident of where it was sampled from.

This separation decides which image the policy sees as its reference. Cold-start SFT shows the original human design, since that stage only establishes the prompt-and-image-to-code format. RL and the benchmark instead show a reference that an image generator produces from the inferred prompt. This is the input the method is built for: at deployment, an image generator turns a design intent into one concrete visual proposal, and a code model implements that proposal as markup. A generated reference supplies a uniform contemporary register, but it is one sample from a generator and need not agree with its prompt in every string or ordering. Training and evaluating on human designs would tune the policy to an input it never receives at deployment.

We collect approximately 18{,}000 human-designed slides, posters, dashboards, and webpages, extract CLIP image features, and cluster the pool. Distances from each cluster center are divided into equal-width intervals, and each sampling round draws equal numbers of samples from these intervals without replacement, so that peripheral designs are covered as well as those near the centers. Strata too sparse to support the allocation are excluded. Because sources are drawn without replacement, the SFT, RL, and benchmark splits are disjoint; \cref{appendix:dataset_examples} checks this directly on the released files.

Most of the pool comes from public design collections released under the MIT and Apache-2.0 licenses; the remainder is internal data used for training only. We do not release the source images. We release the artifacts our pipeline produces: the inferred prompts, the verifiable requirements, the generated reference images, and the generated markup and renderings of every evaluated arm. Since the references are generated from text prompts rather than copied from the source designs, the release contains everything needed to reproduce the RL setup and the benchmark without redistributing the underlying collections.

\subsection{Prompt Inference and SFT Demonstrations}

GPT-5.4-mini infers a generation prompt from each sampled image. The prompt specifies the artifact type and its content, including required text, charts, and data, and names the major layout zones and the reading flow, since a request that lists content without saying how it is organized does not describe a deliverable. It may carry one coarse register cue, such as a warm or cool palette, but prescribes none of the appearance choices the visual channels supervise: no specific colors, component shapes, spacing, or typography. The image supplies that guidance. The requirement record extracted alongside the prompt is stricter and admits only exact strings, chart data, and table data. We call such a prompt \emph{content-only}.

To build SFT data, GPT-5.5 receives the prompt together with the original human design and writes three HTML candidates. We render and score them with our reward system and keep the highest-scoring candidate if its reward exceeds $0.85$; clusters left below their allocation are resampled. This yields 3{,}000 demonstrations, each storing the prompt, the original image, the selected code, and the requirements. No image generator is called at this stage.

\begin{figure}[t]
	\centering
	\includegraphics[width=\linewidth]{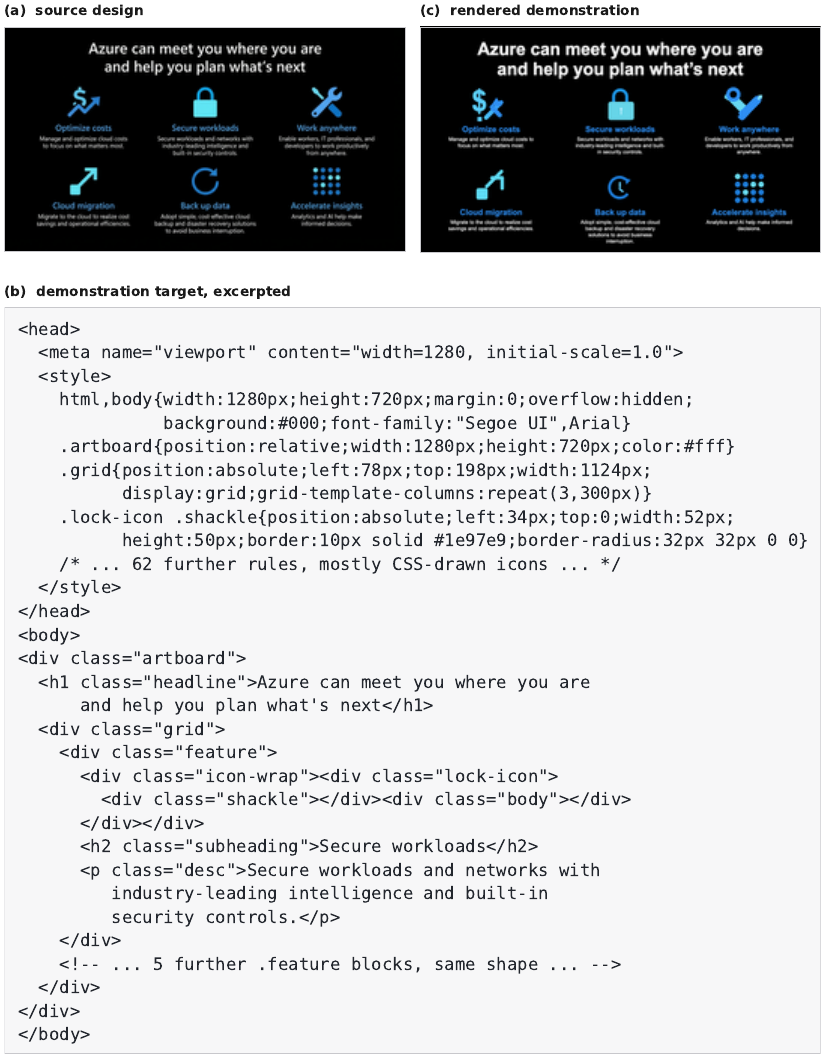}
	\caption{One cold-start SFT demonstration: (a) source design, (b) excerpt of the target HTML with elisions marked, (c) its render in the evaluation sandbox.}
	\label{fig:sft_example}
\end{figure}

\Cref{fig:sft_example} shows one demonstration. Because the supervision target is markup, a demonstration is a triple rather than a pair: the source design the prompt was inferred from, the markup the loss is computed on, and the render the reward system scored during filtering. The excerpt also shows the properties the rule checks depend on. The page is a fixed artboard at the requested canvas size, every visible string lives in its own text node, and icons are built from styled elements rather than fetched as assets, so a render can be read back and checked against the item's requirements.

\subsection{RL and Benchmark References}
\label{appendix:benchmark}

For the pool remaining after SFT construction, we infer content-only prompts and select the RL set by difficulty. The cold-start policy generates eight rollouts per prompt, and the reward system of \cref{sec:appendix_distance_illustration} scores each one, giving a pass@1 and a pass@8 rate per prompt without additional annotation. Prompts solved on nearly every rollout carry little gradient signal under a group-normalized objective, and prompts failed on every rollout carry none, so we keep the band in which pass@1 and pass@8 separate. The retained prompts are then rebalanced across CLIP clusters so that the difficulty filter does not concentrate the RL set in whichever artifact type the SFT policy found hardest. This yields $7{,}408$ RL prompts, shared by both backbones. GPT-Image-2 generates a new reference from each prompt, and the original human design is never shown to the policy.

The held-out benchmark contains 300 items drawn across the four artifact types and the CLIP clusters, with no source design shared with SFT or RL. Prompts and references are constructed exactly as for RL, and the original designs are not used at test time. Each item is stored independently of any model output and contains the natural-language request, format and canvas metadata, atomic verifiable requirements, the prompt-derived evaluation checklist, and the reference.

All systems receive the same prompt and reference wherever their interface supports image conditioning. We generate three rollouts per prompt and average their scores without selection. Rendering failures remain in the denominator, and Execution is reported only as a diagnostic, excluded from Rule, Visual, and Overall. The non-conditional experiments in \cref{tab:noncond} use the same items and scoring and differ only in whether $I_{\mathrm{ref}}$ is shown to the policy.

\subsection{Dataset Statistics}
\label{appendix:dataset_examples}

\Cref{tab:split_stats} summarizes the three splits. Since sampling is stratified over $25$ CLIP clusters rather than over a declared artifact label, the benchmark is balanced by cluster, with $9$ to $20$ items per cluster. Canvas shape is inherited from the source design. A benchmark item carries a median of $12$ verifiable requirements (range $3$ to $23$); the count follows how much legible, checkable content the source contains, because the extraction template prefers fewer high-confidence requirements over broader but noisier ones. We verify split disjointness directly: the $300$ benchmark identifiers have an empty intersection with the RL training file. Twenty-seven benchmark items were drawn from the RL candidate pool and are labeled as such in the release, but they were removed from that pool when drawn and do not appear in the training file.

\begin{figure*}[t]
	\centering
	\includegraphics[width=\linewidth]{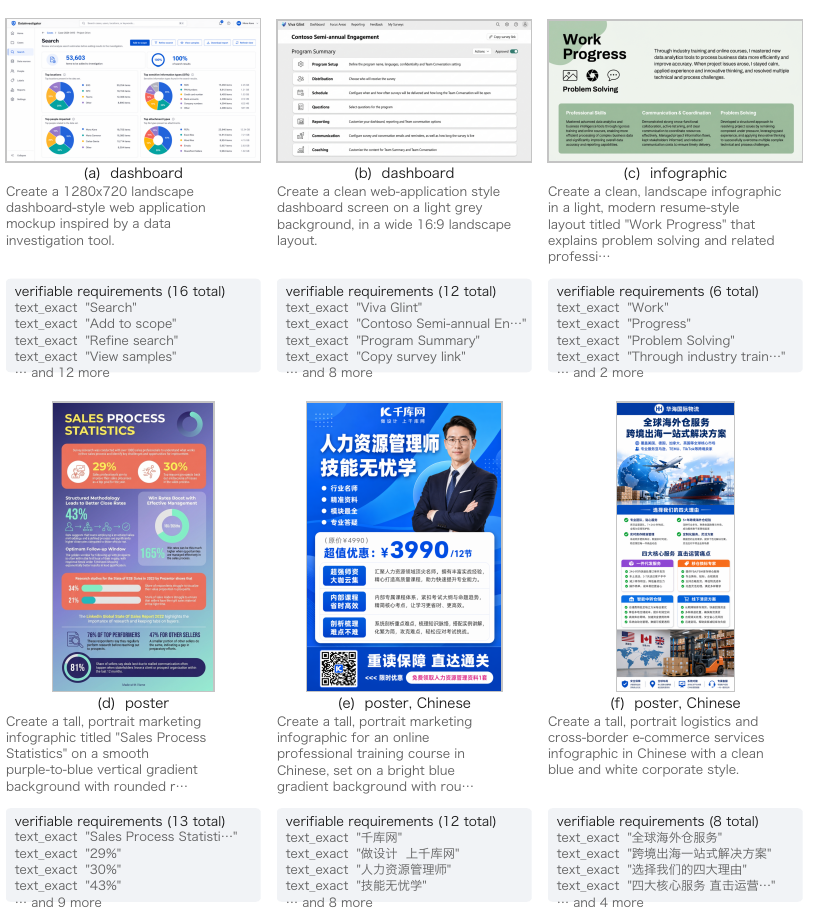}
	\caption{Six benchmark items, each with its GPT-Image-2 reference, the opening of its generation prompt, and its first few verifiable requirements.}
	\label{fig:dataset_examples}
\end{figure*}

\Cref{fig:dataset_examples} shows six benchmark items. What an item asks for is fixed by its prompt and requirements, while the reference supplies visual guidance and is never the object of a string-level check. The items are multilingual and mix canvas shapes, both inherited from the source pool.

\begin{table}[t]
	\centering
	\caption{Composition of the three splits. Both backbones use the same RL file.}
	\label{tab:split_stats}
	\renewcommand{\arraystretch}{1.15}
	\setlength{\tabcolsep}{6pt}
	\begin{tabular}{l r l}
		\toprule
		\textbf{Split} & \textbf{Items} & \textbf{Reference image} \\
		\midrule
		Cold-start SFT         & $3{,}000$            & original human design \\
		RL, 8B and 32B         & $7{,}408$            & GPT-Image-2 from the prompt \\
		Benchmark              & $300$                & GPT-Image-2 from the prompt \\
		\midrule
		\multicolumn{3}{l}{\emph{Benchmark composition}} \\
		CLIP clusters covered  & $25$                 & $9$ to $20$ items each \\
		Orientation            & \multicolumn{2}{l}{$235$ landscape, $56$ portrait, $9$ square} \\
		Requirements per item  & \multicolumn{2}{l}{median $12$, range $3$ to $23$, mean $11.8$} \\
		Requirement types      & \multicolumn{2}{l}{$3{,}299$ \texttt{text\_exact}, $193$ \texttt{chart\_data}, $34$ \texttt{table\_data}} \\
		\bottomrule
	\end{tabular}
\end{table}

\subsection{Prompts Used for Data Construction}
\label{appendix:data_prompts}

Two model calls stand between a source design and a usable training item. The first reads the source image and returns the generation prompt together with its verifiable requirements; the second turns that prompt into a reference image. We print the first template in full, since it carries the contract the reward depends on, and describe the second in prose.

The first call produces the prompt and the requirements in a single JSON object rather than extracting requirements from the prompt afterwards, and it forbids any \texttt{text\_exact} requirement whose string does not appear exactly in the generation prompt. A requirement is thus a marked verification point inside the prompt, not an independent claim about the source image, and a model that follows the prompt cannot be penalized for a string it was never asked to produce. The template keeps style at the level of one coarse category cue, which leaves palette, typography, and decoration to the reference image. Two further rules exist because the target is code: charts must be requested as ECharts with literal data arrays, since otherwise a static picture of a chart would be an acceptable reading of the prompt, and content HTML cannot draw, such as photographs and brand logos, must be requested as a sized \texttt{placehold.co} slot.

\begin{promptbox}[Prompt and requirement inference from a source design, abridged]
\begin{lstlisting}[style=json]
SYSTEM: You write assessment-oriented image-generation prompts by analysing one
real reference infographic. Given one reference image, return exactly one JSON
object and nothing else. Preserve the meaningful information load and evaluation
value of the reference; do not demand a pixel-identical recreation. Use the
image to infer layout, information density, reading flow, major visible text,
charts, and tables. For visual style capture ONLY a coarse category-level cue
(cool/warm palette, serif/sans-serif, flat/illustrative); do not describe
specific colors, font families, or decorative detail. Omit illegible details and
incidental clutter. Do not guess corrupted text or partially legible OCR
fragments; if a string is uncertain, omit it rather than hallucinating it.
Every text_exact requirement must be backed by the generation_prompt itself: if
exact visible text matters, include that exact string in the generation_prompt
before extracting it as a requirement. Only use text_exact, chart_data, and
table_data.
CHARTS: when the image contains any chart, the generation_prompt MUST instruct
the generator to render it as an ECharts chart with literal data arrays.
NON-REPRODUCIBLE VISUALS: the downstream generator emits HTML/CSS only and
cannot render photographs, complex illustrations, hand-drawn art, or brand
logos. When such content appears, the generation_prompt MUST request a
https://placehold.co/WxH?text=Description placeholder of appropriate size.
Simple geometric icons and flat CSS shapes may be described for reproduction.

USER: Write an image-generation prompt for later evaluation based on the
reference image. Return one JSON object matching the schema.
  [Reference Metadata] id, reference image size, aspect ratio, orientation
  1) generation_prompt is one coherent prompt body, not a checklist dump, and
     must not exceed 3000 characters.
  2) Preserve the important information mix: major layout zones, key headings,
     charts, tables, quantitative blocks. Include only one short coarse style
     cue.
  3) Use text_exact for the most important clearly legible strings worth
     evaluating. Every text_exact string must appear verbatim inside
     generation_prompt; if it is not written there, do not include it. Cap at 12
     and calibrate to information density. Use chart_data only for real charts
     whose literal values should remain in ECharts data, and table_data only for
     real visible tables. Do not create structured requirements for layout,
     order, grouping, repeated structure, or count.
  4) Tell the generator to expose correctness through separate visible text
     nodes, natural DOM structure, literal ECharts data arrays, and real HTML
     tables.
  5) - 6) omitted here: output length and formatting boilerplate.
  7) soft_cues is a single short sentence of coarse style cues only.
  [Output Rules] exactly the keys id, reference_metadata, generation_prompt,
  verifiable_requirements, soft_cues. Prefer fewer high-confidence requirements
  over broader but noisy extraction.
  [Schema Contract] {schema_contract}

INPUT: source_image
OUTPUT: {"id": ..., "reference_metadata": ..., "generation_prompt": ...,
         "verifiable_requirements": [...], "soft_cues": ...}
\end{lstlisting}
\end{promptbox}

The second call generates the reference image with GPT-Image-2. The prompt cannot be passed through unchanged, because it was written for a code generator: a \texttt{placehold.co} slot tells the model to leave a region empty, and constraints such as CSS-only icons or no photographs exist only because HTML cannot draw them. Passing these through would make the reference inherit the limitations of the code target. A deterministic rewrite therefore turns each placeholder request into a description of the illustration it stands for and drops the code-only constraints and pixel sizing, while leaving layout structure, text, and chart and table values untouched, so that the reference and the requirements still describe the same artifact. A fixed prefix then asks for a contemporary editorial register and fully drawn illustrations. This instruction applies to the reference only: the code prompt still requests placeholder slots, and the judge counts a placeholder as the image being present. The canvas follows the aspect ratio of the source design, rounded to a multiple of $16$.

\section{Implementation and Additional Results}
\label{appendix:supplementary_results}

This section gives the settings needed to reproduce a run and the results that support the main text. All scores use the Rule and Visual columns of \cref{tab:model_performance}.

\subsection{Implementation Details}
\label{appendix:implementation}
\label{appendix:rendering}

\paragraph{Training configuration.}
\method uses Qwen3-VL-8B and Qwen3-VL-32B backbones. Cold-start SFT uses the 3{,}000 demonstrations above with a learning rate of $10^{-5}$, followed by GDPO on the same $7{,}408$ RL prompts for both scales, so that the two backbones differ only in capacity. We use verl's FSDP--vLLM hybrid engine without a critic or a separately trained reward model. RL uses AdamW with a constant learning rate of $5\times10^{-6}$, no warmup, and weight decay $0.01$. Each step samples 128 prompts with eight rollouts each and runs one optimization epoch; the 8B and 32B models are trained for 400 and 520 steps. KL regularization is applied in the loss rather than the reward, with KL and entropy coefficients both $0.001$. Rollouts use temperature $1.0$ and top-$p$ $1.0$, with prompt and response each capped at 8{,}192 tokens.

\paragraph{Inference and decoding.}
Benchmark generation uses three rollouts per prompt with temperature $0.8$, top-$p$ $0.95$, at most 12{,}000 output tokens, and a model context of 24{,}576 tokens. Closed-source baselines are queried through their hosted APIs with the same prompt and rollout count. Reasoning-mode endpoints do not accept sampling parameters and run at provider defaults; the other hosted arms use $0.8$ and $0.95$. No arm is tuned on our benchmark. All generations were collected between 28 August and 18 September 2026, and we release the generated markup and renderings of every arm.

\paragraph{Rendering sandbox.}
Candidate HTML is rendered from a local file in a sandboxed Playwright browser. The canvas size comes from the reference metadata, with $1280\times720$ used only when it is absent; the viewport is enlarged when necessary and the screenshot is clipped to the canvas. ECharts and placeholder assets are served locally, and other external requests are blocked. The renderer exports the DOM, computed styles, bounding boxes, console status, and the screenshot, and a 120-second safeguard recycles stalled workers.

\subsection{Learning Dynamics of the Decoupled Reward Channels}
\label{sec:analysis_rft_dynamics}

\rewarddynamicsfloat

\Cref{fig:post_training_dynamics} shows that Execution saturates after cold-start SFT, while Text and Chart improve fastest (pass@1 from $\approx\!0.79$ to $\approx\!0.94$ and from $\approx\!0.58$ to $\approx\!0.87$). The design channels improve more slowly and are still rising when training ends, consistent with spatial and perceptual organization being harder to acquire. Boundary stabilizes early with low within-group variance, precisely the kind of signal that scalar aggregation would suppress. Pass@1 converges toward pass@8, so RL makes behavior that previously appeared only in the best rollouts typical of a single sample, and the close training and held-out curves show no visible reward trade-off or overfitting.

\subsection{Loss of Verifiable Capabilities under Reconstruction Objectives}
\label{appendix:reconstruction}

A natural alternative to the decoupled reward is to score a rollout by how closely its render reproduces the reference. \Cref{tab:reconstruction} evaluates this alternative under the same checklist, columns, and aggregation as \cref{tab:model_performance}; the lower block fixes the backbone, training prompts, and sampler and varies only the reward.

A reconstruction scalar does move the visual dimensions it can see: CLIP-similarity GRPO raises Layout from $71.19$ to $78.45$ and Visual from $60.11$ to $65.69$, and an absolute VLM score moves them in the same direction by a smaller margin. It does not preserve capabilities that are discrete and checkable. Chart falls from $42.35$ on the backbone to $9.67$, $9.78$, and $15.27$ across the three reconstruction arms, and the cause is visible in the markup. Among the $89$ prompts with a \texttt{chart\_data} or \texttt{table\_data} requirement, we count a prompt as charted if any of its rollouts calls \texttt{echarts.init}, since loading the library without instantiating a chart yields exactly the static substitute at issue. The backbone charts $73$ prompts and \method-8B $80$, whereas the reconstruction arms chart only $15$, $8$, and $26$, replacing the rest with static markup that is close in pixels and wrong in content. Published image-to-code systems show the same pattern, with UI2Code$^n$ at $7.56$ Chart despite a Visual score of $59.92$. A similarity scalar cannot separate a real chart from a picture of one.

\begin{table}[t]
	\centering
	\caption{\textbf{Reconstruction objectives.} Upper: published image-to-code systems. Lower: the backbone, prompts, and sampler of \method-8B with the reward replaced by a reconstruction objective. Evaluation as in \cref{tab:model_performance}; all rows receive $I_{\mathrm{ref}}$; scores in \%.}
	\label{tab:reconstruction}
	\renewcommand{\arraystretch}{1.16}
	\setlength{\tabcolsep}{7.5pt}
	\resizebox{\linewidth}{!}{%
	\begin{tabular}{l@{\hspace{0.8em}}|@{\hspace{0.8em}} c c c >{\columncolor{groupBasicCol}}c c c c >{\columncolor{groupDesignCol}}c >{\columncolor{oursRow}}c}
		\toprule
		\textbf{Model}
			& \textbf{Text} & \textbf{Bound} & \textbf{Chart} & \textbf{Rule}
			& \textbf{Content} & \textbf{Layout} & \textbf{Style} & \textbf{Visual} & \textbf{Overall} \\
		\midrule
		\footnotesize\textit{Image-to-code systems} & \multicolumn{9}{c}{} \\[-2pt]
		Design2Code           & 39.60 & 54.13 & 0.92 & \cellcolor{groupBasicCol}31.55 & 19.26 & 9.75 & 2.79 & \cellcolor{groupDesignCol}10.60 & 21.08 \\
		UI2Code$^n$           & 80.10 & 62.38 & 7.56 & \cellcolor{groupBasicCol}50.01 & 69.70 & 68.11 & 41.93 & \cellcolor{groupDesignCol}59.92 & 54.96 \\
		WebSight              & 25.28 & 62.48 & 0.06 & \cellcolor{groupBasicCol}29.27 & 27.46 & 20.56 & 7.43 & \cellcolor{groupDesignCol}18.48 & 23.88 \\
		\midrule
		\footnotesize\textit{Reconstruction objective, our data, Qwen3-VL-8B} & \multicolumn{9}{c}{} \\[-2pt]
		Qwen3-VL-8B \textit{(backbone)} & 64.07 & 70.99 & 42.35 & \cellcolor{groupBasicCol}59.14 & 52.90 & 51.33 & 29.92 & \cellcolor{groupDesignCol}44.72 & 51.93 \\
		\quad Reconstruction SFT        & 79.77 & 79.62 & 9.67 & \cellcolor{groupBasicCol}56.35 & 65.96 & 71.19 & 43.20 & \cellcolor{groupDesignCol}60.11 & 58.23 \\
		\quad + CLIP-similarity GRPO    & 84.49 & 86.36 & 9.78 & \cellcolor{groupBasicCol}60.21 & 70.42 & 78.45 & 48.20 & \cellcolor{groupDesignCol}65.69 & 62.95 \\
		\quad + absolute VLM GRPO       & 82.64 & 86.37 & 15.27 & \cellcolor{groupBasicCol}61.43 & 69.27 & 76.03 & 44.95 & \cellcolor{groupDesignCol}63.42 & 62.42 \\
		\addlinespace[1pt]
		\rowcolor{oursRow} \method-8B \textit{(decoupled reward)} & \textbf{94.06} & \textbf{88.36} & \textbf{87.79} & \cellcolor{groupBasicCol}\textbf{90.07} & \textbf{86.41} & \textbf{87.80} & \textbf{53.21} & \cellcolor{groupDesignCol}\textbf{75.80} & \cellcolor{oursRow}\textbf{82.94} \\
		\bottomrule
	\end{tabular}
	}
\end{table}
\subsection{Additional Qualitative Results}
\label{appendix:additional_figures}
\label{appendix:qualitative}

\begingroup
\renewcommand{\topfraction}{0.92}
\renewcommand{\bottomfraction}{0.92}
\renewcommand{\textfraction}{0.06}
\renewcommand{\floatpagefraction}{0.92}

\Cref{fig:gallery_b,fig:gallery_a} show a random sample of \method-32B outputs, complementing the side-by-side comparison in \cref{fig:qualitative_cases} at a size where the pages can be read. Each panel is a single rollout on a randomly drawn held-out benchmark prompt, rendered in the evaluation sandbox and shown unedited. The sample covers dashboards with live charts, diagrams, slides, and tall posters, in several languages, with no per-format change to the prompt or decoding.

\begin{figure}[t]
	\centering
	\includegraphics[width=\linewidth]{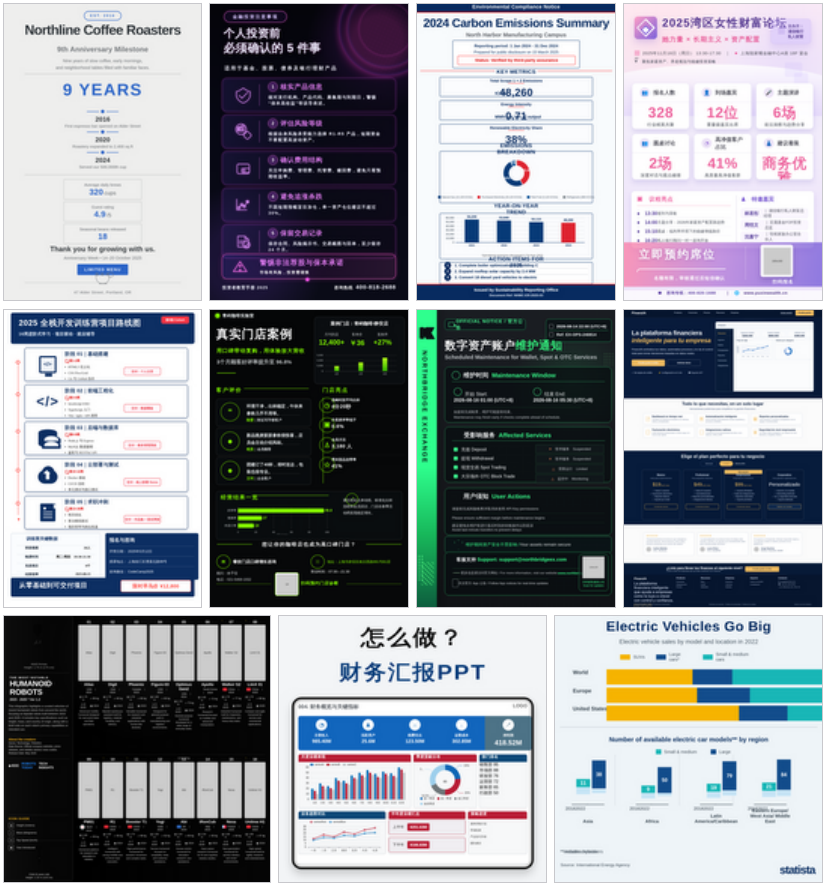}
	\caption{Portrait and square outputs of \method-32B on held-out benchmark prompts.}
	\label{fig:gallery_b}
\end{figure}

\begin{figure}[t]
	\centering
	\includegraphics[width=\linewidth]{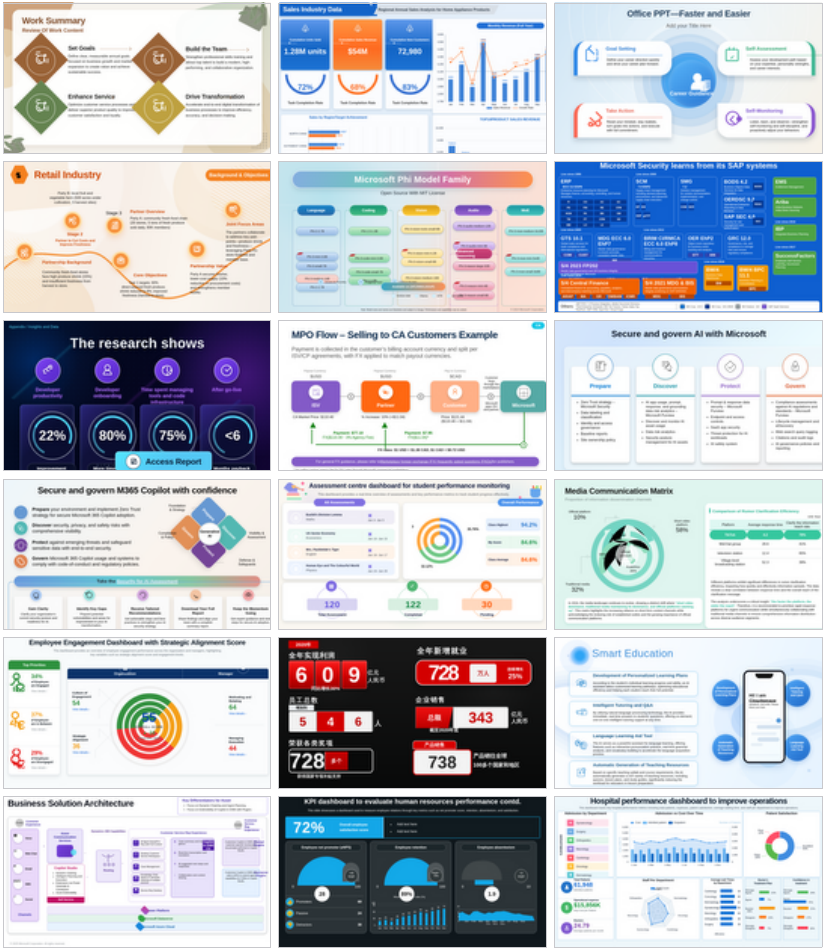}
	\caption{Landscape outputs of \method-32B on held-out benchmark prompts.}
	\label{fig:gallery_a}
\end{figure}

\endgroup

\section{Reward System}
\label{sec:appendix_distance_illustration}

\begin{figure}[t]
	\centering
	\includegraphics[width=\linewidth]{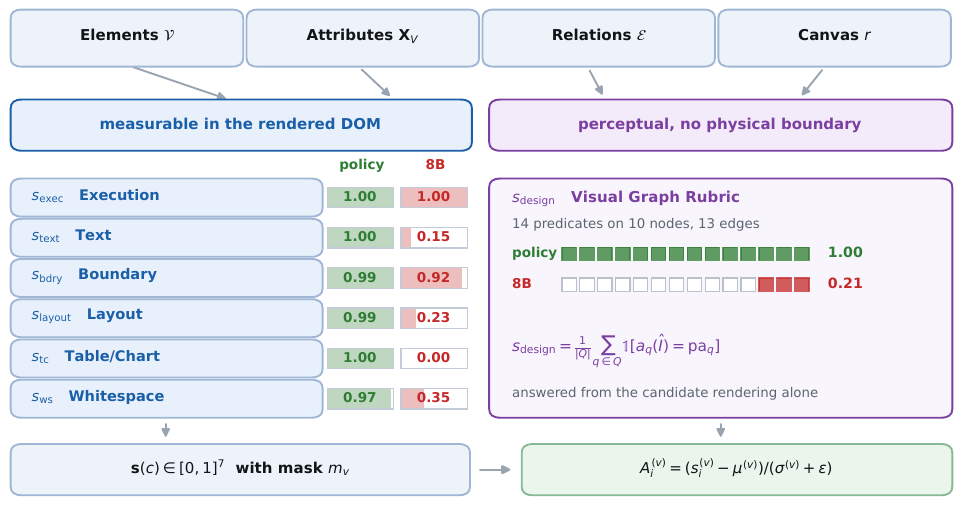}
	\caption{\textbf{Reward channels.} Six deterministic verifiers read the rendered DOM and screenshot. \emph{Execution}: render success, decayed by uncaught JavaScript errors. \emph{Text}: required strings found, weighted by overlap and in-canvas visibility. \emph{Boundary}: in-canvas fraction of overflowing elements and aspect-ratio match. \emph{Layout}: proximity of text blocks matched to the reference. \emph{Table/Chart}: parsed table rows and ECharts series against the required data. \emph{Whitespace}: cosine similarity of occupancy grids with the reference. \emph{VGR}: pass rate of sample-specific yes/no predicates answered from the candidate render. Each channel is masked by applicability and normalized separately by GDPO. Values are from two rollouts on one prompt.}
	\label{fig:reward_mechanism_appendix}
\end{figure}

This section specifies each training reward of \cref{sec:distance}. Reference-side evidence comes from the prompt, its requirements, and $I_{\mathrm{ref}}$; element correspondences use text, type, semantic role, and relative geometry, and no verifier requires pixel-perfect reconstruction. \Cref{fig:reward_mechanism_appendix} summarizes the seven channels. These rewards supply the learning signal only; the reported scores are defined separately in \cref{appendix:evaluation_protocol}.

\subsection{Per-Rollout Reward Record}
\label{appendix:reward_record}

Every rollout produces a structured record rather than a single scalar. For each channel $v$ the record holds an applicability mask $m_v\in\{0,1\}$, a raw score $s_v\in[0,1]$, and a diagnostic payload $e_v$ (\cref{tab:reward_record_schema}). Results marked \textsc{not applicable} are excluded from aggregation, whereas explicit zero-score failures are included. Infrastructure failures are recorded separately from semantic ones, so a judge timeout is never read as negative design feedback, and the same record serves training curves, benchmark aggregation, and qualitative diagnosis.

\begin{table}[t]
	\centering
	\caption{Fields of the per-rollout reward record.}
	\label{tab:reward_record_schema}
	\renewcommand{\arraystretch}{1.15}
	\setlength{\tabcolsep}{5pt}
	\resizebox{\linewidth}{!}{%
	\begin{tabular}{l l l l}
		\toprule
		\textbf{Channel} & \textbf{Applicability} & \textbf{Recorded evidence} & \textbf{Output} \\
		\midrule
		Execution & all rollouts & render status, timeout, blank-page and console flags & pass / severity score \\
		Text & required text exists & required, matched, missing, and unsupported strings & normalized fidelity \\
		Boundary & rendered DOM exists & element and canvas boxes, overflow events, aspect ratios & containment and shape score \\
		TableChart & table or chart required & parsed cells or ECharts type, categories, series, and values & structural accuracy \\
		SemanticLayout & matched text blocks exist & block matches, normalized centers, distances, and coverage & spatial proximity \\
		Whitespace & usable screenshots exist & coarse/fine occupancy grids and per-cell differences & multi-scale similarity \\
		VGR & valid rubric exists & question, binding, pass answer, model answer, and pass flag & predicate pass rate \\
		\bottomrule
	\end{tabular}%
	}
\end{table}

\subsection{Verifiable Requirements}
\label{appendix:requirements}

Verifiable requirements are extracted from the prompt before generation and do not depend on any candidate output. The schema has three types: \texttt{text\_exact} retains the exact required string, and \texttt{chart\_data} and \texttt{table\_data} retain the expected object type, labels, series, and values. Canvas metadata supplies the target dimensions used by Boundary. Pairwise and group relations are not part of this schema; graph-bound VGR predicates supervise them during training, and the prompt-derived Visual checklist evaluates them at test time.

\subsection{Deterministic Verifiers}

\paragraph{Execution.}
Playwright aborts external requests outside the allowlist, and the static resource audit records violations without a direct penalty. Empty HTML and render blockers receive zero. A render is blank when its grayscale variance is below $8.0$ and its dark-pixel ratio below $0.005$, and renders exceeding $15{,}000$ ms also fail. After network and instrumentation errors are filtered, $N_{\mathrm{fatal}}$ uncaught JavaScript errors give $s_{\mathrm{exec}}=\exp(-0.4N_{\mathrm{fatal}})$; other warnings are kept in the payload without penalty. A blocking execution failure short-circuits downstream verification.

\paragraph{Text.}
Text is extracted from the candidate HTML with \texttt{script} and \texttt{style} nodes removed; no reference OCR enters this verifier. Each required string is checked by case-sensitive substring presence after NFKC and punctuation normalization. The binary match is multiplied by two factors computed from the rendered text boxes: the ratio of their union to their summed area, which penalizes overlap, and their mean visible fraction inside the canvas.

\paragraph{Boundary.}
Boundary combines two sub-scores computed from browser bounding boxes. The containment term is the fraction of overflowing element area that still falls inside the canvas,
\begin{equation}
	S_{\mathrm{ovf}}
	=\frac{\sum_{e\in\mathcal O}\operatorname{area}(e\cap C)}{\sum_{e\in\mathcal O}\operatorname{area}(e)},
	\label{eq:appendix_boundary_overflow}
\end{equation}
where $C$ is the declared canvas and $\mathcal O$ collects elements whose horizontal or vertical excess exceeds $8$ or $24$ pixels; hidden nodes, decorative elements, and SVG primitives are excluded. The shape term compares the rendered aspect ratio $r_{\hat I}$ with the declared $r_{\mathrm{ref}}$,
\begin{equation}
	S_{\mathrm{size}}=\exp\!\left(-\lambda\,\frac{\lvert r_{\hat I}-r_{\mathrm{ref}}\rvert}{r_{\mathrm{ref}}}\right),
	\qquad \lambda=2.5,
	\label{eq:appendix_boundary_size}
\end{equation}
and the channel is their mean, $s_{\mathrm{bdry}}=\tfrac{1}{2}(S_{\mathrm{ovf}}+S_{\mathrm{size}})$. The two terms capture unrelated failures: a page can match the canvas shape while spilling a table past its lower edge, or keep every element inside a canvas of the wrong proportions.

\begin{figure}[t]
	\centering
	\includegraphics[width=\linewidth]{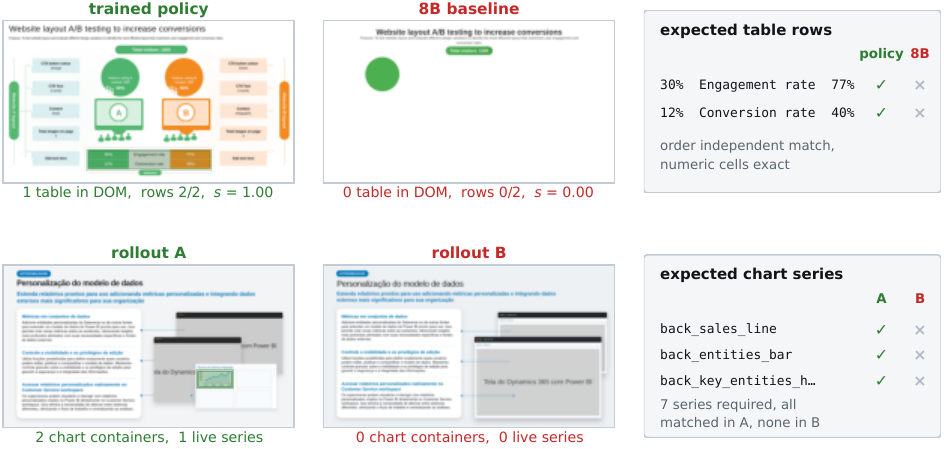}
	\caption{\textbf{Table/Chart.} Outlines mark the tables and charts found in the DOM. Top: the baseline emits no table. Bottom: two rollouts on a chart prompt, with and without live ECharts instances.}
	\label{fig:reward_tablechart}
\end{figure}

\paragraph{Table and chart.}
Tables are parsed from HTML, and each expected row is greedily matched to an unused candidate row regardless of order. Numeric cells require normalized equality, while other cells may match by normalized alphanumeric containment; the table score is the fraction of expected rows reproduced. Charts are read from live ECharts series, falling back to static script extraction. When a chart subtype is required, data and subtype contribute $60\%$ and $40\%$. Both branches are multiplied by the mean in-canvas visible fraction of the elements. The record stores field-level matches, so a correct chart type with wrong values can be told apart from a missing chart (\cref{fig:reward_tablechart}). The channel is \textsc{not applicable} when the prompt requires neither a table nor a chart.

\begin{figure}[t]
	\centering
	\includegraphics[width=\linewidth]{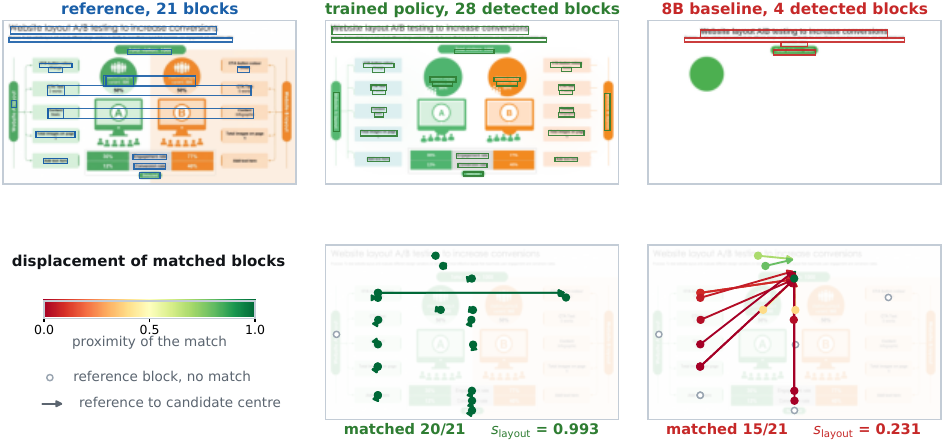}
	\caption{\textbf{SemanticLayout.} Top: OCR text blocks. Bottom: arrows from each reference block to its matched candidate block, coloured by proximity; hollow circles are unmatched reference blocks.}
	\label{fig:layout_reward_comparison}
\end{figure}

\paragraph{Semantic layout.}
SemanticLayout compares where content lands rather than how it is styled. PaddleOCR extracts text boxes from the reference and candidate renders, with candidate DOM boxes as a fallback, and adjacent fragments are merged. For each reference block, candidates with SequenceMatcher similarity below $0.3$ are discarded and the spatially closest remaining one is selected. A matched pair $j$ with centre distance $d_j$ in normalized canvas units receives
\begin{equation}
	\operatorname{prox}_j=\exp\!\left(-\frac{d_j^{2}}{2\sigma^{2}}\right),
	\qquad \sigma=0.15 .
	\label{eq:appendix_layout_prox}
\end{equation}
Writing $\mathcal M$ for the matched pairs and $\mathcal V_{\mathrm{ref}}$ for the merged reference blocks, so that $\rho=\lvert\mathcal M\rvert/\lvert\mathcal V_{\mathrm{ref}}\rvert$ is the match rate, the channel is
\begin{equation}
	s_{\mathrm{layout}}
	=\left(\frac{1}{\lvert\mathcal M\rvert}\sum_{j\in\mathcal M}\operatorname{prox}_j\right)
	\bigl(0.9+0.1\,\rho\bigr).
	\label{eq:appendix_layout_score}
\end{equation}
Colour, typeface, and exact box extent do not enter the score, so a candidate may diverge stylistically but not spatially (\cref{fig:layout_reward_comparison}).

\begin{figure}[t]
	\centering
	\includegraphics[width=\linewidth]{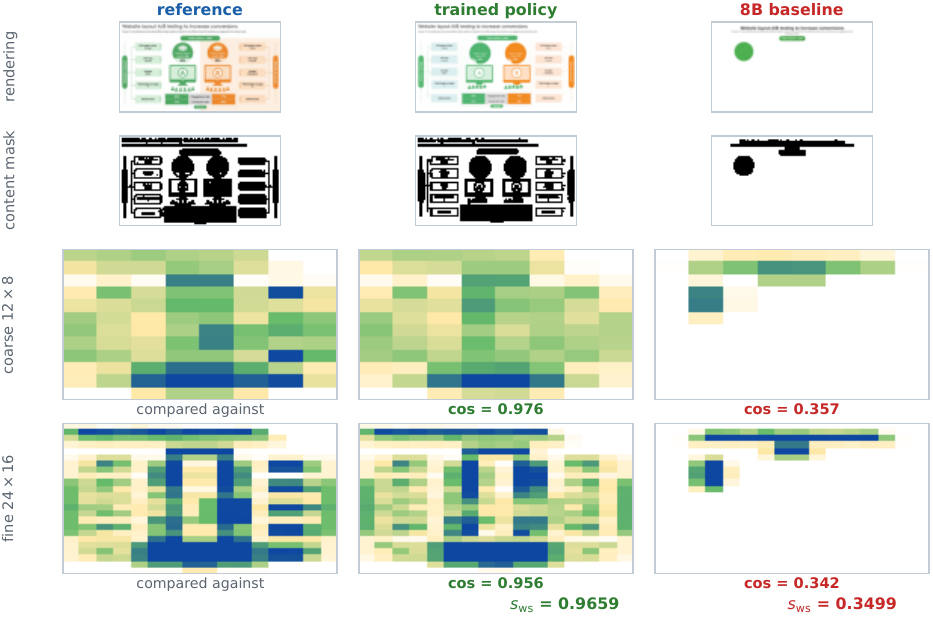}
	\caption{\textbf{Whitespace.} Rendering, content mask, and coarse and fine occupancy grids; $s_{\mathrm{ws}}$ averages the two grid cosines with the reference.}
	\label{fig:reward_whitespace}
\end{figure}

\paragraph{Whitespace.}
We extract a content mask from each screenshot by combining Canny edges with pixels whose intensity differs by more than $25$ from the median border intensity, and rasterize it into coarse ($12\times8$) and fine ($24\times16$) occupancy grids. Each image is gridded in its own pixel space, so a different render size is not itself penalized. With flattened grids $\mathbf g^c$ and $\mathbf g^f$,
\begin{equation}
	s_{\mathrm{ws}}
	=\tfrac{1}{2}\cos\!\left(\mathbf g^c_{\hat I},\mathbf g^c_{I_{\mathrm{ref}}}\right)
	+\tfrac{1}{2}\cos\!\left(\mathbf g^f_{\hat I},\mathbf g^f_{I_{\mathrm{ref}}}\right).
	\label{eq:appendix_whitespace}
\end{equation}
The coarse grid captures global visual mass, and the fine grid remains sensitive to local crowding and empty regions (\cref{fig:reward_whitespace}).

\subsection{Visual Graph Rubric}
\label{appendix:vgr}

GPT-5.5 turns each reference into a rubric in two calls. The first reads the generation prompt and the reference and returns a layout class together with a scene graph of nodes and edges. The second receives that graph with the same prompt and image and writes yes/no questions, each bound to a node, an edge, a set of nodes, or the whole canvas, and typed as structure, completeness, visual attribute, relation, subgraph, forbidden, or global. Fixing the graph before the questions are written keeps the structure a question refers to from drifting, and lets every question cite the element it tests. Textual fidelity is excluded, since the Text verifier handles it. At scoring time GPT-5.5 answers the fixed rubric from the candidate rendering alone, and the reward is
\begin{equation}
	s_{\mathrm{vgr}}
	=\frac{1}{|Q|}\sum_{q\in Q}
	\mathbb 1\!\left[a_q(\hat I)=\operatorname{pa}_q\right].
	\label{eq:appendix_vgr}
\end{equation}
Each stored predicate keeps its type, binding, question, pass-answer polarity, model answer, and pass flag, so we can see which criterion changes between two rollouts even when their aggregate scores are similar.

Three rules of the templates carry most of the behavior we rely on. First, the layout class (free form, symmetric grid, timeline sequence, hub and spoke, or hierarchical) is committed before any question is written and sets the subgraph quota, because a forced subgraph question on a flat grid is only a row-alignment question under a graph label. Second, repeated identical elements are folded into one group node with a count, so four KPI cards produce one layout predicate rather than four existence checks. Third, opaque image and icon slots are addressed only by their geometry and position, never by their depicted content, because the reference generator may draw a different picture in the same slot and a content question would penalize a difference the prompt never specified. A programmatic linter checks every rubric for missing bindings, inconsistent polarity, and questions about text or image content; over the $7{,}408$ training rubrics it flags $20$ ($0.3\%$), and $99.6\%$ of questions bind to an element of their own graph. The template asks for $10$ to $20$ questions, and realized rubrics average $14.1$ (range $12$--$19$).

\begin{promptbox}[Call 1, scene graph extraction, abridged]
\begin{lstlisting}[style=json]
SYSTEM: You are a scene-graph extractor for page designs. Read the brief the
design was produced from and the rendered reference, commit to a layout class
before anything else, then describe the page as a graph. Describe only what is
structurally present. Do not write questions; a later step does that.

R1 BOUNDARIES: a node is a region the eye can separate from its surroundings by
   an edge, a background fill, or an unambiguous gap. Do not invent nodes for
   internal padding, and do not merge two elements that a reader would treat as
   separate cards.
R4 ATTRIBUTES: record only properties that are stable under redrawing, i.e.
   position, extent, aspect ratio, alignment, count, fill and border colour,
   shadow, and chart or icon kind. Do not record the depicted subject of a
   photo, illustration, icon, or logo; use image_kind and the slot geometry.
R6 GROUP FOLDING: 3+ identical elements become ONE group node with attrs.count=N,
   not N separate nodes.

STEP 1  SKELETON + LAYOUT CLASS. Choose one of symmetric_grid | hierarchical |
   timeline_sequence | hub_spoke | free_form. List 5-7 essential elements;
   essential means removing it breaks the story.
STEP 2  GRAPH.
   NODE_TYPES  containers: region, card, panel, list, group
               atoms:      text, chart{chart_type}, table, icon{icon_kind},
                           image{image_kind}, button, input, divider, connector,
                           badge, pyramid, funnel, timeline, banner
   EDGE_TYPES  spatial:  contains, left_of, right_of, above, below,
                         aligned_with{axis}, evenly_spaced{members}
               semantic: connects{direction}, timeline_step{order}, parent_of
   Every node carries a stable id, a type, and an attrs object. Emit a semantic
   edge only where the design actually asserts the link, e.g. a drawn arrow, a
   numbered step, or containment under a parent heading.

INPUT: generation_prompt, reference_image
OUTPUT (strict JSON): {"skeleton": [...], "layout_class": ...,
  "graph": {"nodes": [{"id": ..., "type": ..., "attrs": {...}}, ...],
            "edges": [{"id": ..., "type": ..., "from": ..., "to": ...}, ...]}}
\end{lstlisting}
\end{promptbox}

\begin{promptbox}[Call 2, rubric authoring from the graph, abridged]
\begin{lstlisting}[style=json]
SYSTEM: You are a scene-graph-structured rubric author. You are given the brief,
the rendered reference, and the scene graph already extracted from it. Write
yes/no questions against that graph. Do not add, rename, or re-type graph
elements. Adapt the subgraph quota to the given layout class; forced fake
subgraph questions on flat symmetric layouts are worse than no subgraph at all.
R7 is MANDATORY: every semantic edge in the graph must be exercised by a
matching relation question.

R2 OPAQUE SLOTS: never make the depicted subject of a photo, illustration, icon,
   or logo the subject of a question; the reference generator may legitimately
   draw something else in the same slot.
     BAD:  "Does the portrait depict a smiling person?"
     BAD:  "Is the airplane fully visible?"
     GOOD: "Is there a photo-shaped placeholder slot in the upper-left, above
            the light-gray information panel?"
     GOOD: "Is the upper-left photo region rectangular and approximately 1:1 to
            4:3 in aspect ratio?"
   A depicted-subject noun may appear only as an anchor naming which slot.
R3 POLARITY: every question carries pass_answer: "yes" | "no".
R5 BINDING: every question binds to node:<id> | edge:<id> | subgraph:<id,...> |
   global:page. The id must already exist in the given graph.
R7 SEMANTIC EDGE COVERAGE (mandatory): >=1 connects edge requires a
   semantic_connect question; >=2 timeline_step edges require a
   semantic_sequence question; >=2 parent_of edges require a semantic_hierarchy
   question.
R8 NO TEXT OR IMAGE CONTENT: do not ask whether a string is present or what it
   says, and do not ask what an image shows; both are scored elsewhere.

RUBRIC, 10-20 questions, quotas by layout class:
   structure     3-4   arrangement / split / alignment / row / column
   completeness  2-3   intact / no overlap / fully visible
   visual_attr   1-2   color / shadow / border
   relation      1-3   binary edge predicates; subtype required, one of
                       spatial | semantic_connect | semantic_sequence |
                       semantic_hierarchy
   forbidden     1-2   at least one, e.g. "Are there extra X beyond N?" pass=no
   global        2     exactly two, last, pass=yes: no unintended overlap or
                       occlusion; all content fully contained and unclipped
   subgraph            symmetric_grid / free_form: 0-1, do not force;
                       hierarchical, timeline_sequence, hub_spoke: 1-2.
                       A subgraph question must require graph structure, e.g.
                       "Do the 4 timeline nodes follow strict order
                       2018->2020->2022->2024 via the connector arrows?"
                       and not "Are the 6 cards in a 3x2 grid?"
ATOMICITY: one question = one binary check, no AND/OR.

INPUT: generation_prompt, reference_image, graph
OUTPUT (strict JSON): {
  "rubric": [{"id": ..., "binds_to": ..., "qtype": ..., "subtype": ...,
              "question": ..., "pass_answer": ..., "rationale": ...}],
  "qtype_count": {...}}
\end{lstlisting}
\end{promptbox}

\begin{promptbox}[VGR judging template]
\begin{lstlisting}[style=json]
SYSTEM: You are a strict but fair visual judge.

You will see a SINGLE GENERATED IMAGE and a numbered list of yes/no questions
about it. For EACH question, output a JSON line
{"id": "...", "answer": "yes"|"no"|"unclear"}.

Rules:
- Answer ONLY based on what is actually visible in the generated image.
- "unclear" only if the relevant element is genuinely ambiguous or not visible
  at all (use sparingly).
- Do not assume. If the question asks about a "blue-to-pink gradient" and you
  see a flat blue, answer "no".
- Do not be lenient. The point is to find layout differences from the reference.

Output STRICT JSON (no fences):
{"judgments": [{"id":"q1","answer":"yes"}, {"id":"q2","answer":"no"}, ...]}

USER: <generated image>, then

Judge the following GENERATED image against {n} yes/no questions.

Questions:
  q1. ...
  q2. ...

For each question, answer yes / no / unclear based ONLY on what is visible in
the image.
Output strict JSON: {"judgments": [{"id":"q1","answer":"yes"}, ...]}
\end{lstlisting}
\end{promptbox}

The judge answers the whole list in one call, which keeps the cost of the channel independent of rubric length. It sees only the candidate rendering and the questions, never the reference: the reference enters the reward only through the questions authored from it, which keeps the channel from degenerating into an image-similarity score. Two complete rubrics follow.

\providecommand{\vgrexamplewidth}{0.86\linewidth}
\providecommand{\vgrportraitwidth}{0.8\linewidth}
\paragraph{Two complete rubrics.}
\label{appendix:vgr_example}
These examples illustrate the training VGR on benchmark instances, using the same generation prompts as the final evaluation. All archived questions are retained; reference-derived appearance and spatial criteria are not instructions in the generation prompts. Pass denotes the required answer, not a judgment of the reference. \Cref{fig:vgr_example_input,fig:vgr_portrait_input} show the two inputs.
\begin{figure}[!ht]
\noindent\textbf{Product features and connectors (15 questions).}\par\medskip
\centering
\includegraphics[width=\vgrexamplewidth]{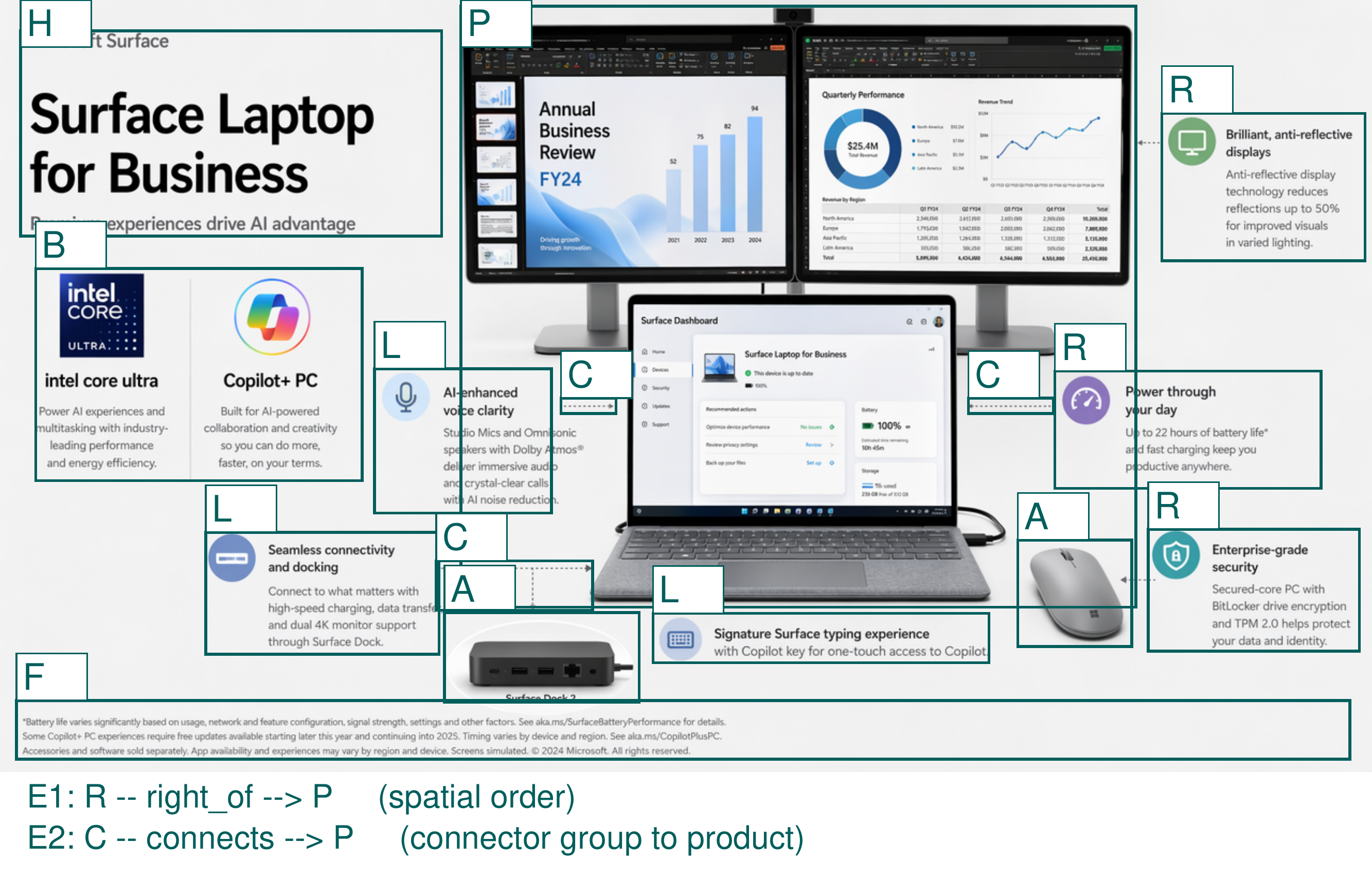}
\par\medskip
\begin{minipage}{\linewidth}
\small\textbf{Complete generation prompt}\par\smallskip
Create a landscape technology marketing infographic promoting a business laptop.\par\smallskip
Include an introductory section with the branding text ``Microsoft Surface'', the main title ``Surface Laptop for Business'', and the subtitle ``Premium experiences drive AI advantage''.\par\smallskip
Follow with two key feature callouts:
- ``intel core ultra'', with concise descriptive text about AI-enabled collaboration, productivity, and on-device AI performance.
- ``Copilot+ PC'', with concise descriptive text about AI-enabled collaboration, productivity, and on-device AI performance.\par\smallskip
Include a workspace setup section showing the business laptop experience with:
- dual external monitors with business apps
- a Surface Laptop dashboard
- a docking device
- a wireless mouse\par\smallskip
Include four to six short supporting feature blurbs covering:
- connectivity and docking, including high-speed charging, data transfer, and dual 4K monitor support
- audio, microphones, and AI-enhanced voice clarity
- ``Anti-reflective display technology reduces reflections up to 50\% for improved visuals in varied lighting''
- security features such as drive encryption and TPM protection
- ``Signature Surface typing experience with Copilot key for one-touch access to Copilot''\par\smallskip
At the end, include a fine-print legal or availability note written like a typical disclaimer paragraph.
\end{minipage}
\caption{Complete generation prompt and reference for the VGR example (instance \texttt{sesimic\_folder\_2\_original\_14504}). Both inputs match the frozen final-evaluation instance. Reference-derived design criteria appear in the VGR questions, not in the generation prompt.}
\label{fig:vgr_example_input}
\end{figure}
\par\smallskip
\noindent Boxes locate stored nodes or groups; their coordinates are manual display annotations, not recorded geometry. Repeated labels mark members of the same group. C marks visible connector regions, not a claim that every target connector is present. E1 and E2 retain stored relation types and endpoints. Canvas denotes the full image.
\begin{center}\footnotesize
\begin{tabular}{ll@{\qquad}ll}
\hline
ID & Stored node & ID & Stored node \\
\midrule
H & \texttt{n\_headline\_block} & B & \texttt{n\_logo\_tile\_group} \\
P & \texttt{n\_hero\_product\_slot} & A & \texttt{n\_accessory\_slot\_group} \\
L & \texttt{n\_left\_callout\_group} & R & \texttt{n\_right\_callout\_group} \\
C & \texttt{n\_connector\_group} & F & \texttt{n\_footer} \\
Canvas & \texttt{global:page} & & \\
\midrule
E1 & \multicolumn{3}{l}{\texttt{edge:e\_right\_callouts\_right\_of\_hero}: R right of P} \\
E2 & \multicolumn{3}{l}{\texttt{edge:e\_callouts\_connect\_to\_product}: C connects to P} \\
\bottomrule
\end{tabular}\end{center}
\FloatBarrier
\begingroup\small\setlength{\tabcolsep}{3pt}\renewcommand{\arraystretch}{1.12}
\begin{longtable}{@{}p{0.06\linewidth}p{0.15\linewidth}p{0.68\linewidth}r@{}}
\toprule ID & Target & Question & Pass \\\midrule\endhead
\midrule\multicolumn{4}{r}{\footnotesize Continued on next page}\endfoot
\bottomrule\endlastfoot
q1 & H & Is the large headline block positioned in the upper-left quadrant of the poster? & Yes \\
q2 & P & Is the central product image slot the largest visual element on the page? & Yes \\
q3 & B & Are the two badge/logo-like tiles arranged as a horizontal pair below the headline area? & Yes \\
q4 & A & Are the two accessory image slots placed along the lower half of the poster rather than near the top? & Yes \\
q5 & P & Is the central product image slot fully visible without being cropped by the page edges? & Yes \\
q6 & L & Are the three left-side callout blocks readable as separate blocks with clear spacing between them? & Yes \\
q7 & R & Are the three right-side callout blocks kept inside the page boundary? & Yes \\
q8 & C & Are the connector lines rendered as thin dotted or dashed gray lines? & Yes \\
q9 & R & Do the right-side callouts use colored circular icon badges next to their text blocks? & Yes \\
q10 & E1 & Is the right-side callout group positioned to the right of the central product image slot? & Yes \\
q11 & E2 & Do visible dotted connectors link the surrounding feature callouts toward the central product or accessory slots? & Yes \\
q12 & F & Is the fine-print footer placed below the main product and accessory area? & Yes \\
q13 & R & Are there extra right-side feature callouts beyond the expected three? & No \\
q14 & Canvas & Are all major visible elements free of unintended overlap or occlusion? & Yes \\
q15 & Canvas & Are all content elements fully contained, unclipped, and unfragmented? & Yes \\
\end{longtable}
\endgroup
\par\medskip
\noindent\textbf{Portrait FAQ grid and footer relations (14 questions).}\par\smallskip
\noindent In \cref{fig:vgr_portrait_input}, B locates the clinic badge; D, the header divider; G, the six-card FAQ grid; C, the contact strip; I, its three icons; F, the footer region; W, the warning block; and Q, the QR placeholder. Canvas denotes the whole page. Boxes are display annotations, not recorded geometry.
\begin{center}\footnotesize
\begin{tabular}{ll@{\qquad}ll}
\hline
ID & Stored node & ID & Stored node \\
\midrule
B & \texttt{n\_top\_badge} & D & \texttt{n\_header\_divider} \\
G & \texttt{n\_faq\_cards} & C & \texttt{n\_contact\_strip} \\
I & \texttt{n\_contact\_icons} & F & \texttt{n\_bottom\_area} \\
W & \texttt{n\_warning\_block} & Q & \texttt{n\_qr\_slot} \\
Canvas & \multicolumn{3}{l}{\texttt{n\_page} / \texttt{global:page}} \\
\midrule
E1 & \multicolumn{3}{l}{\texttt{edge:e\_divider\_above\_cards}: D above G} \\
E2 & \multicolumn{3}{l}{\texttt{edge:e\_warning\_left\_of\_qr}: W left of Q} \\
\bottomrule
\end{tabular}\end{center}
\begingroup\small\setlength{\tabcolsep}{3pt}\renewcommand{\arraystretch}{1.12}
\begin{longtable}{@{}p{0.06\linewidth}p{0.15\linewidth}p{0.68\linewidth}r@{}}
\toprule ID & Target & Question & Pass \\\midrule\endhead
\midrule\multicolumn{4}{r}{\footnotesize Continued on next page}\endfoot
\bottomrule\endlastfoot
q1 & Canvas & Is the overall composition a vertical poster with all major content stacked from top to bottom? & Yes \\
q2 & G & Are the six FAQ cards arranged as a 3-column by 2-row grid? & Yes \\
q3 & C & Is the contact information strip a single wide rounded panel spanning most of the poster width? & Yes \\
q4 & F & Does the bottom area use a two-part layout with the warning block on the left and the QR-code slot on the right? & Yes \\
q5 & G & Is each FAQ card fully visible within its rounded card boundary? & Yes \\
q6 & Q & Is the QR-code placeholder area fully visible as a square slot in the lower-right section? & Yes \\
q7 & I & Are the three contact icons contained inside the contact strip? & Yes \\
q8 & B & Does the top-left badge have a thin neon purple-pink rounded outline? & Yes \\
q9 & G & Do the FAQ cards use dark translucent fills with purple outlined rounded borders? & Yes \\
q10 & E1 & Is the thin horizontal divider positioned above the FAQ card grid? & Yes \\
q11 & E2 & Is the warning block positioned to the left of the QR-code placeholder area? & Yes \\
q12 & G & Are there extra FAQ cards beyond the six-card grid? & No \\
q13 & Canvas & Are all major visible elements free of unintended overlap or occlusion? & Yes \\
q14 & Canvas & Are all content elements fully contained, unclipped, and unfragmented? & Yes \\
\end{longtable}
\endgroup
\begin{figure}[!ht]
\centering
\begin{minipage}{\vgrportraitwidth}
\begin{minipage}[t]{0.45\linewidth}
\vspace{0pt}
\includegraphics[width=\linewidth]{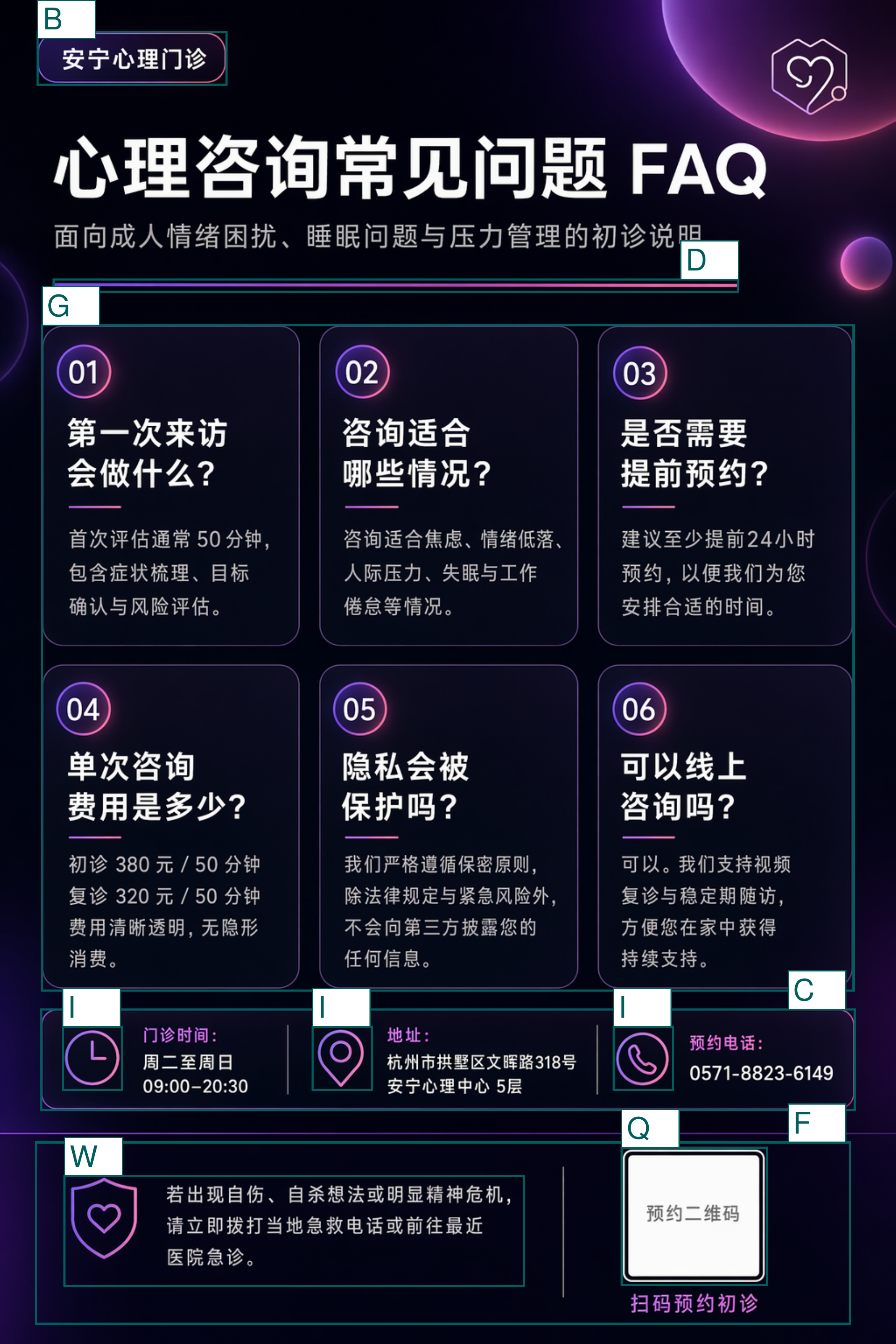}
\end{minipage}\hfill
\begin{minipage}[t]{0.53\linewidth}
\vspace{0pt}
\small\textbf{Complete generation prompt}\par\smallskip
\includegraphics[page=1,width=\linewidth]{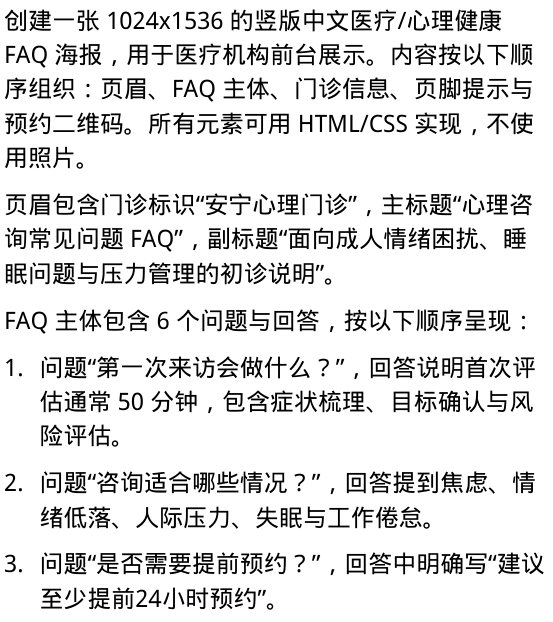}
\end{minipage}
\par\medskip
\includegraphics[page=2,width=\linewidth]{figures/rubric_examples/portrait_plain_prompt.pdf}
\end{minipage}
\caption{Portrait FAQ example (instance \texttt{poster\_v2\_00018}). The complete Chinese generation prompt and reference match the frozen evaluation inputs. The reference is shown once, with display annotations locating the stored groups and relations.}
\label{fig:vgr_portrait_input}
\end{figure}

\FloatBarrier

\section{Evaluation Protocol and Human Study}
\label{appendix:evaluation_protocol}
\label{appendix:final_evaluation}

The final benchmark does not reuse the reference-conditioned SemanticLayout, Whitespace, or VGR rewards of \cref{sec:appendix_distance_illustration}; every reported score is computed from the prompt and the candidate rendering alone. Rule averages the prompt-derived Text, Boundary, and Table/Chart scores, and $89$ of the $300$ prompts carry at least one table or chart requirement. Visual is produced by the checklist pipeline below. Because both Visual and the training rubric rest on model judgments, the section closes with a blinded human study that audits them.

\subsection{Visual Checklist}
\label{appendix:visual_checklist}

Checklist questions are authored from the prompt and its parsed requirements, and proposals that are ambiguous, not decidable from a rendering, or that prescribe a particular implementation are discarded. The pipeline was developed on a separate 50-instance set that does not overlap the benchmark, then frozen and applied unchanged to the $300$ benchmark items. The frozen checklists contain $10{,}727$ questions, of which $1{,}925$ are Content, $4{,}565$ Layout, and $4{,}237$ Style, with $26$--$50$ questions per prompt (mean $35.8$). Each question is answered from the candidate rendering, a dimension score is the pass rate of its questions, and Visual is the mean of the available dimensions; the two prompts with no Content question have that dimension excluded rather than scored zero.

Every question names the element it is about, is decidable from a single rendering, and carries a kind label that maps to one dimension. Content collects kinds that ask whether a required element is present and intact, such as located presence, element count, and element integrity. Layout collects kinds about how elements sit relative to one another or to the canvas, such as collision, containment, structure, and page balance. Style collects kinds about the quality of rendering, such as element quality, icon quality, concrete attribute, and page density. Per-kind quotas keep a page from being scored almost entirely on whether its strings are present. The contract each question must satisfy is given below.

\begin{promptbox}[Checklist question contract, abridged]
\begin{lstlisting}[style=json]
Each question must:
  - be answerable YES or NO from a single rendered image, with no reference
    image and no access to the markup;
  - name the element it is about, so that the answer can be attributed;
  - state any threshold it depends on, such as a fraction of the canvas or a
    ratio between margins;
  - carry an explicit pass_answer, since some questions state a property the
    page should have and others ask whether a defect is present;
  - declare its kind, which fixes the dimension it is reported under.

Reject a proposed question if it:
  - asks about the content of a placeholder image slot;
  - prescribes a particular implementation rather than a visible outcome;
  - restates a browser default that any page satisfies;
  - bundles several claims that could not fail independently.
\end{lstlisting}
\end{promptbox}

A small number of whole-page Style questions are written as disqualifier lists, in which any one of several defects fails the page; the delivery-gate questions in \cref{appendix:eval_rubric_example} are of this form. They ask whether the page clears a publication bar at all, so splitting them would report one verdict several times. They bind to the canvas, and all element-level questions follow the atomicity rule.

At scoring time the frozen checklist is answered one category at a time over the same image, so the judge has a single decision criterion per call. We reproduce the judging template in full. Three of its provisions matter for reading the numbers. The canvas is fixed, so an element that runs past the edge counts as lost content, while full-bleed bands that merely touch the edge are not defects. Placeholder boxes are required by our prompts and count as the image being present, the evaluation-side counterpart of the opaque-slot rule in \cref{appendix:vgr}. Every answer must cite the region it was read from, and an answer that cannot is recorded as \texttt{unclear} and scored as a failure, so a property the judge cannot verify does not silently pass.

\begin{promptbox}[Checklist judging template]
\begin{lstlisting}[style=json]
SYSTEM: You are a strict but fair judge of a generated web page.

You see ONE GENERATED PAGE IMAGE and a numbered list of yes/no questions about
it. There is no reference or target image; you are never asked whether the page
resembles anything.

{brief}                  <- the brief for this call, printed in the next box

Rules:
- Answer ONLY from what is visible in the image.
- The image is a FIXED-SIZE CANVAS, like a slide or a poster. It is not a
  scrolling page and it was not extended to fit the content: anything the page
  laid out beyond this canvas is gone and you are seeing all there is. So an
  element that runs into the canvas edge and stops mid-way -- a severed word, a
  card with no bottom, a row of tiles whose last tile is sliced -- is content
  that was genuinely lost, and it is a real defect. Judge the composition as a
  design that had to fit this canvas.
- Full-bleed design is legitimate, not a defect: background bands, headers,
  footers and images may deliberately extend to the page edge. The test is
  whether something is CUT THROUGH, not whether it touches the edge.
- Read each question's polarity carefully. Some ask whether a DEFECT is present
  or whether a requirement was broken; for those, "yes" means the problem IS
  there.
- Grey placeholder boxes, often labelled with their pixel size, are REQUIRED by
  the brief: the models are told to stand images in with https://placehold.co
  rather than draw them. Treat a placeholder as the image being present and
  correctly placed. Never call it missing, unfinished, low-contrast or a design
  defect.
- Questions quote exact text. Match it by its distinctive words; differences in
  case or punctuation do not matter, but missing or different content does.
- A page carrying more content is not thereby worse, and a page carrying less is
  not thereby better. Do not report a defect unless you can point to the
  specific elements involved; adjacent or tightly spaced is not the same as
  overlapping.
- Do not be lenient and do not be charitable. If a property is only partly true,
  answer what the visible evidence shows.

How to decide each answer. This is a procedure, not advice, and it is the same
for every question and every category:

1. Before you answer, find the evidence in the image and put it in `where`: the
   region you looked at and what you saw there. `where` is REQUIRED on EVERY
   answer. "yes" and "no" are equally claims about the image and both must be
   shown. Neither one is the safe default.
2. If you cannot write that `where` from the image, you do not have the
   evidence, and "unclear" is your only remaining answer. You may not fall back
   on "yes", and you may not fall back on "no" either.
3. Answer "unclear" whenever the evidence is missing, ambiguous, too small to
   resolve, or you would answer differently on a second look. Say in `where`
   what you looked for and could not resolve.

Both answers need evidence because the questions do not share a polarity. Some
state a property the page should have, where "yes" is the good outcome. Others
ask whether a defect is present, where "no" is the good outcome. You are not
told which kind you are looking at, and you must not try to infer it: decide
only what the image shows.

What does NOT count as evidence, in either direction:
- Not seeing something. Failing to find a defect does not establish that a
  stated property holds, and failing to find an element does not establish that
  it is absent. A region you did not resolve is "unclear", not "no".
- The page looking clean, orderly, professional or complete overall.
- What a region probably contains, judged from the rest of the page.
- The brief, the question's own wording, or what a page like this usually does.
- A guess you would not repeat if you looked again.

"unclear" is scored as a FAILURE. That is intended. A question you cannot
resolve should not earn a point, whichever answer would have earned it. Do not
avoid "unclear" to seem decisive, and do not treat it as a last resort: on a
question whose evidence you cannot locate, it is the correct answer, not a
failure of nerve.

Output STRICT JSON, no code fences. `where` must be filled on every judgment:
{"judgments": [
  {"id":"C_title","answer":"yes",
   "where":"title bar, top-left, reads 'Quarterly Review'"},
  {"id":"L_overlap","answer":"no",
   "where":"checked the gap between the chart and the legend at mid-right;
            roughly 20px of clear space, no marks crossing"},
  {"id":"L_grid","answer":"unclear",
   "where":"looked for the fourth card in the lower row; that area is blank
            and I cannot tell if it was dropped or intended"}]}

USER: <generated page image>, then the numbered question list for this call.
\end{lstlisting}
\end{promptbox}

The brief substituted at \texttt{\{brief\}} is the only part of the prompt that changes between calls, and it states the standard of proof for its category. The relation and style briefs demand a higher standard than completeness, because those are the categories where a page that merely looks orderly is easiest to mistake for one that satisfies the question.

\begin{promptbox}[Per-dimension briefs substituted into the judging template]
\begin{lstlisting}[style=json]
completeness:
  Each question asks whether a required element is PRESENT and visible. Judge
  presence only; do not deduct for it looking different from what you might
  expect. Presence means you located the element and can say where it is, not
  that the page has a region where it would belong. Where a question fixes a
  count or a region, both must hold. An element that is present but cut through
  by the canvas edge is content that was lost.

relation:
  Each question asks whether a stated spatial or structural relation between
  elements holds: side by side, stacked, in a row or grid, contained, ordered,
  clear of one another. Answer these to a HIGHER standard of proof. A pass
  requires that you locate BOTH elements named in the question and can see the
  relation between them; a page that merely looks orderly is not evidence. If a
  question asks whether one thing stays clear of another, check the actual gap:
  if their marks touch, cross or print over one another anywhere, that is a
  failure even when the rest of the page is fine. If a question asks whether
  something is contained, the whole element must be inside, not most of it.
  Partly true is not true. Where you cannot find one of the named elements, or
  cannot resolve the relation at this resolution, answer 'unclear' rather than
  assuming the page is fine.

style_strict:
  Each question asks whether ONE stated visual property holds, or whether a
  complex construct such as a timeline, chart, flow or icon is properly realised
  rather than crudely faked. Answer these to a HIGHER standard of proof than the
  other categories. A pass requires that you can point at the specific pixels
  that establish the property, and that you would give the same answer looking
  again. Partly true is not true. A construct that merely gestures at the form,
  such as boxes in a row standing in for a timeline with no connecting spine, or
  an icon that is an unrecognisable blob, does not pass. Where you cannot find
  that specific evidence, answer 'unclear' rather than guessing; 'unclear' is
  scored as a failure, which is the correct outcome for a property you cannot
  actually verify. Do not credit a page for effort, for looking busy, or for
  browser defaults such as a heading already being larger than body text.
\end{lstlisting}
\end{promptbox}

\providecommand{\evalexamplewidth}{0.86\linewidth}
\paragraph{A complete checklist.}
\label{appendix:eval_rubric_example}
This example reproduces all 33 evaluation questions for one instance, covering Content, Layout, and Style. It was selected for legible policy and metric groups, not model scores. \Cref{fig:eval_rubric_example_input} shows the prompt and an image illustrating the subject matter.

\begin{figure}[!ht]
\centering
\includegraphics[width=\evalexamplewidth]{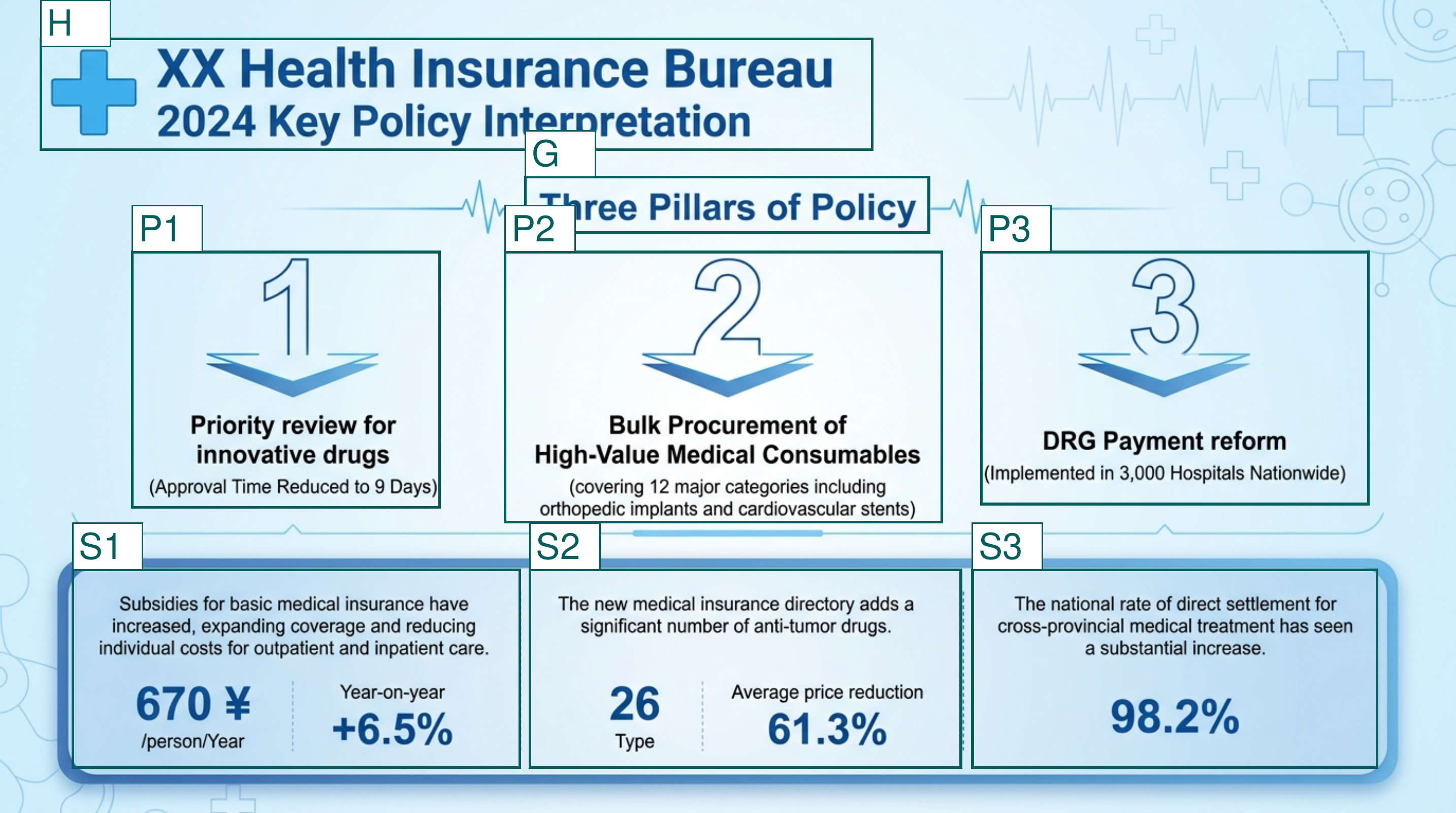}
\par\medskip
\begin{minipage}{\linewidth}
\small\textbf{Complete generation prompt}\par\smallskip
Create a landscape infographic in a wide format about the ``2024 Key Policy Interpretation'' for ``XX Health Insurance Bureau''.\par\smallskip
Begin with the organization name ``XX Health Insurance Bureau'', followed by the subtitle ``2024 Key Policy Interpretation''. Then add the heading ``Three Pillars of Policy''.\par\smallskip
Present three numbered policy pillars in order:\par\smallskip
1. ``Priority review for innovative drugs''  
   Include the supporting text: ``(Approval Time Reduced to 9 Days)''\par\smallskip
2. ``Bulk Procurement of''  
   ``High-Value Medical Consumables''  
   Include the supporting text: ``(covering 12 major categories including orthopedic implants and cardiovascular stents)''\par\smallskip
3. ``DRG Payment reform''  
   Include the supporting text: ``(Implemented in 3,000 Hospitals Nationwide)''\par\smallskip
After the three pillars, include a section with three content groups containing short explanatory text and key metrics.\par\smallskip
The first group should cover subsidies. Include a concise paragraph about subsidies, followed by the metric ``670 \textyen{}'' and the unit ``/person/Year''. Also include the label ``Year-on-year'' with the value ``+6.5\%''.\par\smallskip
The second group should mention that the new medical insurance directory adds anti-tumor drugs. Include two metrics: ``26'' with the label ``Type'', and ``Average price reduction'' with the value ``61.3\%''.\par\smallskip
The third group should describe the national rate of direct settlement for cross-provincial medical treatment. Include the key value ``98.2\%''.\par\smallskip
Do not include charts.
\end{minipage}
\caption{Complete generation prompt and illustrative reference for the evaluation example (instance \texttt{vendor\_data\_original\_903}). The prompt comes from the frozen final-evaluation data.}
\label{fig:eval_rubric_example_input}
\end{figure}
\FloatBarrier
\par\smallskip
\noindent H locates the title block; G, the section heading; P1--P3, the three policy pillars; and S1--S3, the subsidy, drug-directory, and cross-provincial-settlement groups. Canvas means the whole page. These are manual reading aids for this figure, not additional graph bindings or layout requirements in the evaluator. Question IDs, dimension assignments, and pass polarities are retained from the records; ``Pass'' states the required answer and does not report a model outcome.
\begingroup\small\setlength{\tabcolsep}{3pt}\renewcommand{\arraystretch}{1.12}
\begin{longtable}{@{}p{0.06\linewidth}p{0.15\linewidth}p{0.68\linewidth}r@{}}
\multicolumn{4}{@{}l}{\textbf{Content}}\\[4pt]
\toprule ID & Target & Question & Pass \\\midrule\endhead
\midrule\multicolumn{4}{r}{\footnotesize Continued on next page}\endfoot
\bottomrule\endlastfoot
N10 & Canvas & Are there more than zero chart visualizations in the content sections? & No \\
N11 & P1--P3 & Are all three numbered policy pillars drawn whole, with no part of any pillar severed by the canvas edge? & Yes \\
N12 & S3 & Is the cross-provincial direct settlement content, including ``98.2\%'', drawn wholly inside the canvas rather than cut off by a page edge? & Yes \\
N14 & S3 & Are all digits of ``98.2\%'' fully visible, with none clipped by a page edge or by the boundary of whatever element holds the value? & Yes \\
N15 & S1 & Is the subsidies group, including the ``670 \textyen{}'' metric and its unit, drawn whole rather than cut off by a page edge? & Yes \\
\end{longtable}
\endgroup
\begingroup\small\setlength{\tabcolsep}{3pt}\renewcommand{\arraystretch}{1.12}
\begin{longtable}{@{}p{0.06\linewidth}p{0.15\linewidth}p{0.68\linewidth}r@{}}
\multicolumn{4}{@{}l}{\textbf{Layout}}\\[4pt]
\toprule ID & Target & Question & Pass \\\midrule\endhead
\midrule\multicolumn{4}{r}{\footnotesize Continued on next page}\endfoot
\bottomrule\endlastfoot
q1 & H & Is the title block, carrying ``XX Health Insurance Bureau'' and ``2024 Key Policy Interpretation'', encountered before the three numbered policy pillars? & Yes \\
q2 & G / P1--P3 & Does the section heading ``Three Pillars of Policy'' read as introducing the three numbered pillars, rather than attaching to the title block or to the later content groups? & Yes \\
q3 & P1--P3 & Does each numbered policy pillar keep its own supporting text with it? Answer no if any pillar's supporting text sits closer to a different pillar than to its own. & Yes \\
q4 & S1--S3 & Do the three metric content groups read as one later section, encountered after the three numbered pillars rather than interleaved with them? & Yes \\
q6 & S1--S3 & Are all three post-pillar content groups present and legible on the page? & Yes \\
q10 & G / P1--P3 & Is the section heading ``Three Pillars of Policy'' encountered before the first numbered pillar? & Yes \\
q11 & S1--S3 & Does each metric content group keep its own explanatory text and its own metric values together, with neither spilling into a neighbouring group? & Yes \\
X01 & P1--P3 & Are the three numbered policy pillar items presented in the brief's 1--2--3 order, with each number attached to its own pillar title? & Yes \\
X02 & S1--S3 & Do the three metric content groups read as three distinct groups, with each group's metric values visually attached to that group rather than to a neighbour? & Yes \\
N16 & P1--P3 & Do any of the three numbered policy pillars overlap each other? Count only shared area or content crossing into a neighbouring pillar, not a narrow gap. & No \\
N17 & P2 & Does the supporting text of the second policy pillar run outside that pillar into a neighbouring pillar? & No \\
N18 & S3 & Do decorative or background elements overlap the ``98.2\%'' callout? Count only actual overprinting or touching of the digits or their label. & No \\
N19 & S1--S3 & Do dividers, borders, or background shapes cut through the metric numbers or the explanatory text of the three content groups? & No \\
H03 & P2 / S1--S3 & Is the second pillar's parenthetical supporting text kept visually with the second pillar and clear of the later content groups? Answer no if it sits closer to a content group than to its own pillar, or if that supporting text is absent. & Yes \\
H04 & P1--P3 & Do the three numbered pillar titles stay legible, with no other element crossing through or covering their lettering? Answer no if any pillar title is missing or its lettering is obscured enough to be unreadable. & Yes \\
WS01 & Canvas & Consider the page as a whole, ignoring its background colour. Is the content spread over the page, with no empty region larger than about a quarter of the page? Answer no if the content stops partway down and leaves a wide empty band across the bottom, if a whole column or side of the page carries nothing, or if there is a large hole in the middle. Measure the empty region against the full page, not against the neighbouring blocks. Normal separation between sections is not an empty region. If you cannot judge the extent of the empty area, answer unclear. & Yes \\
\end{longtable}
\endgroup
\begingroup\small\setlength{\tabcolsep}{3pt}\renewcommand{\arraystretch}{1.12}
\begin{longtable}{@{}p{0.06\linewidth}p{0.15\linewidth}p{0.68\linewidth}r@{}}
\multicolumn{4}{@{}l}{\textbf{Style}}\\[4pt]
\toprule ID & Target & Question & Pass \\\midrule\endhead
\midrule\multicolumn{4}{r}{\footnotesize Continued on next page}\endfoot
\bottomrule\endlastfoot
H06 & S1 & In the subsidies group, is the unit ``/person/Year'' rendered with the ``670 \textyen{}'' metric so it clearly belongs to that value? Answer no if the unit is not fully readable, or if it appears to label a different metric. & Yes \\
SL01 & Canvas & Is there a visible hierarchy across the page, and do the title block, the section heading ``Three Pillars of Policy'', the three numbered pillars, and the three content groups read as distinct levels rather than one undifferentiated block? Answer no if any of the four reads at the same weight as its neighbours, or if a reader cannot tell where one level ends and the next begins. & Yes \\
SL02 & P1--P3 & Does every numbered policy pillar keep its number clearly associated with its own pillar title, and does the visual reading order of the pillars match 1, 2, 3? Answer no if any number appears closer to a different pillar title, or if the numbers are encountered out of order. & Yes \\
SL03 & Canvas & Do the title block, every numbered policy pillar, and every post-pillar content group use coordinated but distinct visual treatment so the pillars read as one set and the content groups as another? Answer no if any member of either set is given visibly different treatment from its peers. & Yes \\
SL04 & S1--S3 & Do all specified metric values in the three post-pillar content groups read with their own unit or label as clear metric pairings? Answer no if any metric pairing is unreadable or visually attaches to the wrong content group. & Yes \\
SL05 & Canvas & Is every numbered policy pillar marker and every specified metric value visually distinguishable from the surrounding explanatory text? Answer no if any marker or value blends into nearby paragraph text. & Yes \\
SL06 & S1--S3 & Does each of the three post-pillar content groups keep its explanatory text plus numeric values as one clear cluster? Answer no if any group's explanatory text sits closer to another group's metric than to its own, or if a cluster is unreadable from overlap. & Yes \\
\end{longtable}
\endgroup
\subsubsection*{Style: shared judging instruction for DG01--DG05}
{\small Judge this against finished professional design work, the standard of a studio's released output. Do not judge it against the other attempts you have seen, and do not judge it as a hunt for mistakes. A page can contain no outright error and still be a long way from that standard. Being plain, safe, generic or underdeveloped is not neutral here: it means the work is unfinished, and a page with little on it has more design left to do, not less.}\par
The preceding instruction is an identical prefix in each of DG01--DG05; it is printed once here. Each complete question is that prefix followed by its row below. No question text is omitted.
\begingroup\small\setlength{\tabcolsep}{3pt}\renewcommand{\arraystretch}{1.12}
\begin{longtable}{@{}p{0.06\linewidth}p{0.15\linewidth}p{0.68\linewidth}r@{}}
\toprule ID & Target & Question & Pass \\\midrule\endhead
\midrule\multicolumn{4}{r}{\footnotesize Continued on next page}\endfoot
\bottomrule\endlastfoot
DG01 & Canvas & Is the colour work already at that standard, so that a designer taking this over would change nothing about it? Answer no if the palette is a default or a safe choice rather than a considered one, if colour only fills shapes instead of carrying structure and meaning, if the page falls back on plain white and grey because no palette was committed to, if any hue is off, or if the tints and neutrals have not been tuned to sit together. & Yes \\
DG02 & Canvas & Is the typography already at that standard, so that a designer taking this over would change nothing about it? Answer no if the type reads as browser defaults dressed up, if the scale is a few sizes chosen by eye rather than a system, if two sizes do the same job, if the page never commits to a display size where it needs one, or if line length, line spacing or optical alignment would be corrected in review. & Yes \\
DG03 & Canvas & Has the material on this page actually been designed rather than merely placed, so that a designer would neither add anything nor take anything away? Answer no if a section is a plain list or a plain block that a designer would have given form to, if a region is empty because nothing was developed for it rather than because the composition wanted the rest, if anything reads as a placeholder, if filler is occupying space, or if something important was given less room than it deserves. & Yes \\
DG04 & Canvas & Is this page composed, so that the eye is led first, second and third exactly as intended and nothing would be adjusted? Answer no if the page is only stacked in a reasonable order without being composed, if nothing on it is genuinely dominant, if the first thing seen is not the most important thing, if two elements compete for the same rank, or if any part goes flat and stops directing the eye. & Yes \\
DG05 & Canvas & Is every repeated component resolved to the standard of released work, so that none of them would be revisited? Answer no if any card, row, tile, chip, label or control is a default container rather than a designed one, if one instance is less resolved than its neighbours, if spacing inside repeated elements varies without reason, or if a designer picking this up would rebuild the components rather than adjust them. & Yes \\
\end{longtable}
\endgroup

\subsection{External Benchmark Protocols}
\label{appendix:external}

We evaluate Design2Code, SlidesBench, and GDB/lica under their official protocols. Design2Code measures block matching, text similarity, position, color, and CLIP similarity against the target webpage, and generations that fail to render receive zero. SlidesBench scores content, layout, and color on a 0--5 scale. For GDB/lica we report NIMA, CLIP, PickScore, FID, color harmony, and OCR fidelity, where OCR is the fraction of ground-truth tokens recovered by Tesseract (page segmentation mode 6) after the benchmark's normalization; on the same 100 renders it gives 0.719, 0.561, and 0.685 for GPT-5.5, Qwen3-VL-32B, and \method-32B. All rows of \cref{tab:external_benchmarks} are run by us with one decoding, rendering, and scoring pipeline per benchmark, so scores are comparable across systems within a benchmark but not across benchmarks.

\subsection{Human Study on Our Benchmark}
\label{appendix:human_study}

We recruited 52 participants and retained 150 completed questionnaires after quality control; questionnaires completed by the same participant contain disjoint items, and all participants gave informed consent. Each questionnaire has 35--36 formal items, namely six judge-reliability, 10--11 Content, three or four aesthetic, and 16 consistency items, plus an instruction check and an obvious visual sentinel. Every formal item is judged by three different participants. Block and item order are counterbalanced, and system identity, automatic scores, machine answers, and reference images are hidden. Incomplete sessions, failed checks, and image-loading failures are excluded independently of response direction, and uncertainty estimates account for repeated observations within sessions and briefs.

\paragraph{Judge reliability.}
This study asks whether people reproduce the local decisions of the automatic judges. We sample 150 predicates each from the training VGR and the Visual checklist, stratified by dimension. Each page shows one candidate image and one predicate, answered Yes, No, or Unclear, and the automatic answer is compared with the three-person majority. For the automatic judge, Unclear is a failure because every answer must cite evidence; for people, Unclear is missing, since abstaining is not asserting that a property is absent. Agreement is therefore computed over predicates whose human majority is Yes or No.

\paragraph{Content and aesthetic quality.}
This study tests whether \method-32B satisfies more explicit Content requirements than GPT-5.5 while matching its aesthetic quality. The Content task covers all 267 eligible briefs with one frozen image per system: of the 300 briefs, 30 are reserved for the pilot that fixed the protocol, two have no Content question, and one lacks a complete set of GPT-5.5 renders. Participants see the brief, one image, and its Content questions; the two systems appear on separate pages, and no session contains both versions of a brief. Responses are averaged within question and then within image, with Unclear treated as missing. The aesthetic task shows one image from each system for 80 briefs, without the brief or checklist, and asks for a 1--7 rating. Equivalence requires the entire 95\% confidence interval of the difference to lie within the pre-specified $[-0.5,+0.5]$ margin.

\paragraph{Score--preference consistency.}
This study asks whether Visual scores predict human preference. Five images for each of 80 briefs form ten within-brief pairs. Participants see one anonymized pair without the brief and choose A better, B better, Tie, or Unclear; each pair receives three judgments with balanced left--right order. For pairs with unequal scores, choosing the higher-scoring image counts 1, Tie 0.5, and the lower-scoring image 0, with Unclear missing. Equal-score pairs are reported separately, and agreement is broken down by score gap.

\subsection{Human Study on External Benchmarks}

We additionally collect 50 completed questionnaires of 16 or 17 pairwise items for Design2Code, SlidesBench, and GDB/lica. For each benchmark, the eight evaluated systems form 28 model pairs, from each of which we freeze two score-blind, same-task comparisons, giving 56 image pairs per benchmark and 168 overall; every pair is judged five times, for 840 judgments. Each page shows two anonymized outputs and asks which is more aesthetically polished, with the task, reference, system identity, and scores hidden.

Each benchmark is summarized by a per-sample composite. Design2Code averages Block, Text, Position, Color, and CLIP; SlidesBench averages Content, Layout, and Color; GDB/lica averages normalized NIMA, CLIP, PickScore, Color Harmony, OCR, and a reversed paired feature distance, since FID is defined only over a set of images. Agreement is scored as above and averaged first within and then across pairs. To compare gaps across scales we use $100|s_A-s_B|/R$, where $R$ is the metric range, grouped into $<5$, $5$--$15$, and $\ge15$; confidence intervals use a source-task cluster bootstrap (\cref{tab:external_consistency}).

\begin{table}[t]
	\centering
	\caption{Human agreement with each metric's score ordering, by relative score gap.}
	\label{tab:external_consistency}
	\small
	\setlength{\tabcolsep}{5pt}
	\begin{tabular}{lccc}
		\toprule
		\textbf{Metric} & \textbf{$<5\%$ of range} & \textbf{$5$--$15\%$ of range} & \textbf{$\ge15\%$ of range} \\
		\midrule
		Our Visual & 66.72\% & 80.00\%  & 93.76\% \\
		Design2Code & 44.67\% & 66.84\% & 82.50\% \\
		SlidesBench & \textnormal{--} & 74.00\% & 91.25\% \\
		GDB/lica & 52.63\% & 74.67\% & 90.23\% \\
		\bottomrule
	\end{tabular}
\end{table}

\FloatBarrier
\section{Failure Analysis}
\label{appendix:failure_analysis}

\Cref{sec:analysis_failure} names three residual failure modes. This appendix measures them and analyzes why the relational and semantic modes persist. All measurements use the $300$-prompt benchmark and the three judged rollouts of \cref{tab:model_performance}.

\subsection{Measuring the three modes}
\label{appendix:failure_axis}

\begin{figure}[t]
	\centering
	\includegraphics[width=\linewidth]{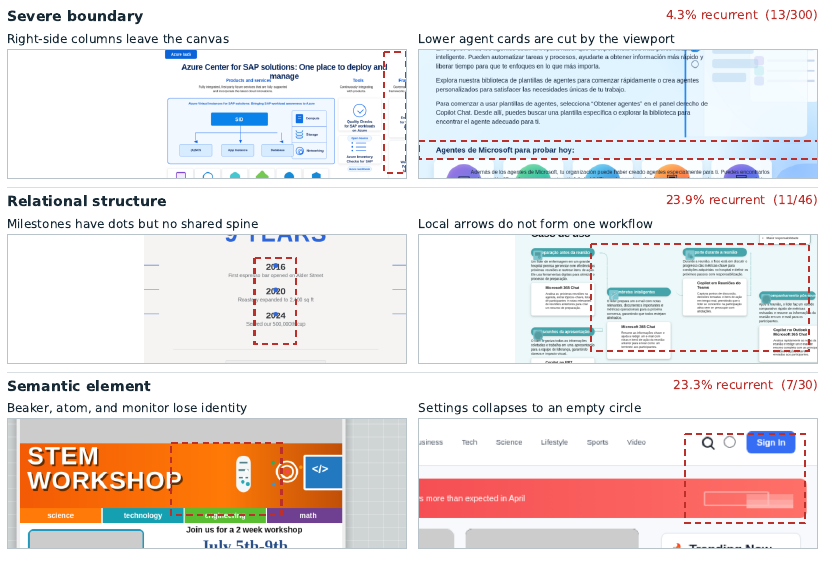}
	\caption{Additional instances of the three residual failure modes, with recurrence rates above each column.}
	\label{fig:failure_cases_appendix}
\end{figure}

A failure is recurrent when it appears in at least two of the three rollouts for a prompt. Boundary applies to every page and is measured over all $300$ prompts. The other two modes apply only to prompts that call for a connected structure or for semantic elements such as icons, so they are measured over the $46$ and $30$ prompts whose checklist contains the corresponding predicate. \Cref{tab:failure_rates} reports the rates, and \cref{fig:failure_cases_appendix} shows two further instances of each mode.

On Boundary, the paired difference to GPT-5.5 is $-30.3$ points ($[-36.0,-24.7]$). The residual cases share a shape. The dashboard in \cref{fig:failure_cases} fails in all three rollouts: its two lower panels begin at $y=535$ px and each request $410$ px of height on a $720$ px canvas. Every region is constructed, but independent fixed-pixel allocations exceed the page height, and the running total is the one quantity that stays invisible until the page is rendered.

\begin{table}[t]
	\centering
	\caption{Residual failure modes. Scope: number of prompts to which the mode applies. Recurrent: failure in $\geq 2$ of $3$ rollouts (lower is better); mean pass: higher is better.}
	\label{tab:failure_rates}
	\small
	\renewcommand{\arraystretch}{1.15}
	\setlength{\tabcolsep}{5pt}
	\begin{tabular}{l l cc}
		\toprule
		\textbf{Mode (scope)} & \textbf{Metric} & \textbf{\method-32B} & \textbf{GPT-5.5} \\
		\midrule
		Severe boundary ($300$) & recurrent & $\mathbf{4.3\%}$ & $34.7\%$ \\
		\midrule
		Relational subgraph ($46$) & recurrent & $23.9\%$ & $\mathbf{10.9\%}$ \\
		& mean pass & $75.4\%$ & $\mathbf{87.0\%}$ \\
		\midrule
		Semantic element ($30$) & recurrent & $23.3\%$ & $\mathbf{3.3\%}$ \\
		& mean pass & $74.3\%$ & $\mathbf{92.0\%}$ \\
		\bottomrule
	\end{tabular}
\end{table}

\subsection{Why the relational and semantic modes persist}
\label{appendix:asymmetry}

Relational structure is supervised explicitly. The design graph carries order, connection, and hierarchy edges such as \texttt{timeline\_step}, \texttt{connects}, and \texttt{parent\_of}, the layout class sets a subgraph quota, and VGR binds relation- and subgraph-type predicates to these edges (\cref{appendix:vgr}). The mode persists because it is the hardest objective to optimize, for two reasons.

First, credit for a relation is implicit. A relation predicate is one of about fourteen rubric questions whose pass rate forms a single channel, and GDPO normalizes per channel rather than per question, so the advantage of a violated relation is shared with every other criterion on the page. Subgraph-level supervision is also sparse and coarse: only $46$ of the $10{,}727$ checklist predicates are subgraph-level, and a failed one marks the page rather than the responsible edge.

Second, the objective is intrinsically demanding. On the $46$ relational prompts, pairwise relation predicates pass at $95.3\%$ and local structure predicates at $95.8\%$, while the whole-subgraph predicate passes at $75.4\%$. Drawing nodes and local arrangements is local and readily learned; holding a timeline, hierarchy, or branch set together as one connected object requires knowledge of the content, reasoning over it, and planning the global layout around it. GPT-5.5 shows the same ordering, so the difficulty lies in the predicate type rather than in our pipeline. The five-link example in \cref{fig:failure_cases} illustrates this: every link element exists, and one shared geometry rule removes every vertical stroke at once.

Semantic elements fail at a similar level of abstraction. In \cref{fig:failure_cases}, the controls sit in plausible slots with a consistent treatment, yet a phone, camera, gear, beaker, atom, or monitor becomes hard to tell apart. Rendering a recognizable icon means mapping a meaning to a drawn form in fine-grained SVG, which draws on the visual knowledge of the backbone. Our rewards supervise the slot's presence, geometry, and treatment, and by design leave its depicted content unscored, because the reference generator may draw a different picture in the same slot (\cref{appendix:vgr}).

\makeatletter
\let\clearpage\ms@clearpage
\makeatother
\clearpage

\end{document}